\documentclass[sigconf]{acmart}

\renewcommand\footnotetextcopyrightpermission[1]{}
\usepackage{enumitem}
\usepackage{tabularx, booktabs}
\usepackage{subcaption}
\usepackage{caption}
\usepackage{longtable}
\usepackage[most]{tcolorbox}
\usepackage{multirow}
\usepackage{fontawesome5}

\setlist[itemize]{noitemsep, topsep=0pt}
\setlist[enumerate]{noitemsep, topsep=0pt}

\AtBeginDocument{}

\copyrightyear{2025} 
\acmYear{2025} 
\acmConference[Agentic and GenAI Eval @KDD25]{KDD workshop on Evaluation and Trustworthiness of Agentic and Generative AI Models}{4 August 2025}{Toronto, ON}

\begin{document}

\title{Reasoning Beyond the Obvious: Evaluating Divergent and Convergent Thinking in LLMs for Financial Scenarios}

\author{Zhuang Qiang Bok}
\authornote{Both authors contributed equally to this research.}
\email{zq.bok@deepinsightlabs.ai}
\affiliation{\institution{Deep Insight Labs} \country{Singapore}}

\author{Watson Wei Khong Chua}
\authornotemark[1]
\email{watson_chua@tech.gov.sg}
\affiliation{\institution{Government Technology Agency} \country{Singapore}}

\begin{abstract}
Most reasoning benchmarks for LLMs emphasize factual accuracy or step-by-step logic. In finance, however, professionals must not only converge on optimal decisions but also generate creative, plausible futures under uncertainty. We introduce \textbf{ConDiFi}, a benchmark that jointly evaluates divergent and convergent thinking in LLMs for financial tasks.

ConDiFi features 607 macro-financial prompts for divergent reasoning and 990 multi-hop adversarial MCQs for convergent reasoning. Using this benchmark, we evaluated 14 leading models and uncovered striking differences. Despite high fluency, GPT-4o underperforms on Novelty and Actionability. In contrast, models like DeepSeek-R1 and Cohere Command R+ rank among the top for generating actionable, insights suitable for investment decisions.

ConDiFi provides a new perspective to assess reasoning capabilities essential to safe and strategic deployment of LLMs in finance.
\end{abstract}



\maketitle

\section{Introduction}

Evaluating the depth and nature of the reasoning behind Large Language Models' (LLMs) outputs is a persistent challenge. Models may appear capable of “reasoning,” but it is often unclear whether their success arises from genuine cognitive processes or from memorizing statistical patterns learned during pretraining on massive datasets \cite{rein2023gpqa}.

“Reasoning” is frequently cited as a core LLM strength, but the term itself is broad and poorly defined. It can refer to anything from simple recall and arithmetic to strategic planning and analogical inference. This ambiguity hinders meaningful evaluation. We argue that reasoning should be disaggregated into precise cognitive constructs to enable more rigorous analysis.

In this work, we focus on two well-established constructs from cognitive psychology: \textit{divergent thinking}—which involves generating multiple, novel possibilities, and \textit{convergent thinking}—which aims at identifying the best solution under constraints. These modes of thinking capture different but complementary dimensions of intelligence: divergent thinking is critical for creativity, ideation, and strategy formation \cite{bellemare2024divergent, kumar2025human, liang2024encouraging}, while convergent thinking is central to tasks requiring logical deduction and precision \cite{kumar2025human}.

This duality is particularly salient in fields like finance, where professionals must devise innovative responses to market shifts and synthesize those ideas into sound, actionable decisions. Whether structuring a novel financial instrument or crafting a risk mitigation strategy, both imaginative exploration and disciplined analysis are essential. Despite this, current evaluations overwhelmingly emphasize convergent tasks, leaving divergent capabilities underexplored—often reduced to narrow tests like the Alternative Uses Task (AUT), which are prone to data contamination when used to benchmark LLMs, and lack real-world relevance.

To address this, we introduce a novel benchmark explicitly designed to assess both divergent and convergent reasoning in the financial domain. Our contributions are: 
\begin{itemize}
    \item Two tailored datasets: one for evaluating structured, logic-driven financial problem-solving, and another for assessing open-ended, creative financial foresight. Both are constructed with data sources dated post-LLM training to minimize contamination, and to test domain-specific understanding and cognitive flexibility.
    \item A comprehensive divergent and convergent reasoning capability evaluation of 14 commonly-used models at the time of writing, across a wide spectrum of model sizes and reasoning optimizations. 
\end{itemize}

Through this benchmark and evaluation, we aim to highlight the limitations of existing reasoning evaluations \cite{golde2025mastermind, su2025bright, wu2024cofca, wu2024mrke}, provide a more holistic and domain-relevant standard for measuring the cognitive capabilities of LLMs in the financial domain.

\section{Related Work}

Recent Large Language Model (LLM) advancements have driven research into nuanced benchmarks for reasoning, creativity, and domain expertise. However, few comprehensively assess both divergent (idea generation) thinking and convergent (solution refinement) in tandem, particularly within high-stakes, structured domains like finance. This is a critical gap, given the need for AI to generate novel insights beyond mere knowledge retrieval in such domains.

Early LLM evaluations (e.g., NQ \cite{kwiatkowski2019natural}, MS MARCO \cite{nguyen2016msmarco}, BEIR \cite{thakur2021beir}) focused on factual recall or general retrieval. While benchmarks like MMLU \cite{hendrycks2021mmlu}, ARC \cite{clark2018arc}, and OpenBookQA \cite{mihaylov2018openbookqa} broadened domain coverage, they, along with more recent efforts probing deeper reasoning like BRIGHT \cite{su2025bright} and TheoremQA \cite{chen2023theoremqa} (which includes finance), predominantly assess convergent thinking and often lack structured financial domain modeling. Advanced benchmarks such as ARB \cite{sawada2023arb}, GPQA \cite{rein2023gpqa}, MASTERMINDEVAL \cite{golde2025mastermind}, and NATURALREASONING \cite{yuan2025naturalreasoning} also generally overlook financial scenario reasoning. Multi-step reasoning benchmarks (e.g., HotpotQA \cite{yang2018hotpotqa}, 2WikiMultihopQA \cite{ho2020constructing}) can suffer from data leakage (e.g., HotpotQA \cite{wu2024mrke, rein2023gpqa}); efforts like MRKE \cite{wu2024mrke} and COFCA \cite{wu2024cofca} use counterfactuals to mitigate this, but their application in finance is limited.

Creativity evaluations distinguish divergent and convergent thinking \cite{kumar2025human}, but classic tests (RAT, AUT) are contaminated for LLMs. Domain-specific approaches include DENIAL PROMPTING for code \cite{lu2024benchmarking} and DSI for writing originality \cite{bellemare2024divergent}. While LLMs can aid creativity, they might hinder independent human divergent thinking \cite{kumar2025human}. Techniques like MAD \cite{liang2024encouraging} foster thought diversity but do not directly assess divergent outputs. These efforts rarely target constrained, high-stakes financial creativity where originality and viability are paramount.

Data contamination from widely available content remains a pervasive issue for many benchmarks. Comprehensive meta-benchmarks like BIG-Bench \cite{srivastava2023imitationgame} and HELM \cite{liang2024encouraging}, while broad, often aggregate reasoning types and offer limited coverage or disaggregation for professional domains like finance. 

Our work addresses these gaps by introducing a financial benchmark that disaggregates divergent (ideation) and convergent (constrained solution selection) reasoning. It uses original, post-training scenarios to mitigate contamination and reveal capabilities hidden by general-purpose tests.

\section{The Convergent-Divergent for Financial (ConDiFi) Reasoning Benchmark}
\subsection{Dataset Overview}

Our dataset comprises 607 scenarios for divergent thinking and 990 questions for convergent thinking.

Each question set in the divergent thinking dataset is a medium-length summary (less than 500 words) of real events (dated 1 May 2025 or after) for a company. Each scenario and summary was picked and written with reference to economics, financial, geopolitical, and political issues of concern to professional investors. Given a scenario, the answering model is prompted to speculate by generating a branching timeline of how the scenario may evolve based on sound economics, financial, and political science methodologies. 

Each question set in the convergent dataset is a Multiple Choice Question (MCQ) with one correct answer and 3 distractors. For each question in the convergent dataset, a scenario is given about a specific company listed in the New York Stock Exchange (NYSE). After which, four different options are given where each option is a sequence of events which might happen after the event. One correct answer satisfies all criteria for convergent thinking (i.e. factor alignment, temporal coherence, logical entailment) while the others are incorrect as they violate at least one criterion. The correctness of each option, as well as its correctness is also given in the dataset. An example is shown in Appendix \ref{appendix:convergent-refinement-prompt-template}.

\subsection{Dataset Construction Pipeline}
\noindent\textbf{Divergent Thinking Dataset}\par\vspace{0.3em}
\begin{figure}[htbp]
    \centering
    \includegraphics[width=\linewidth]{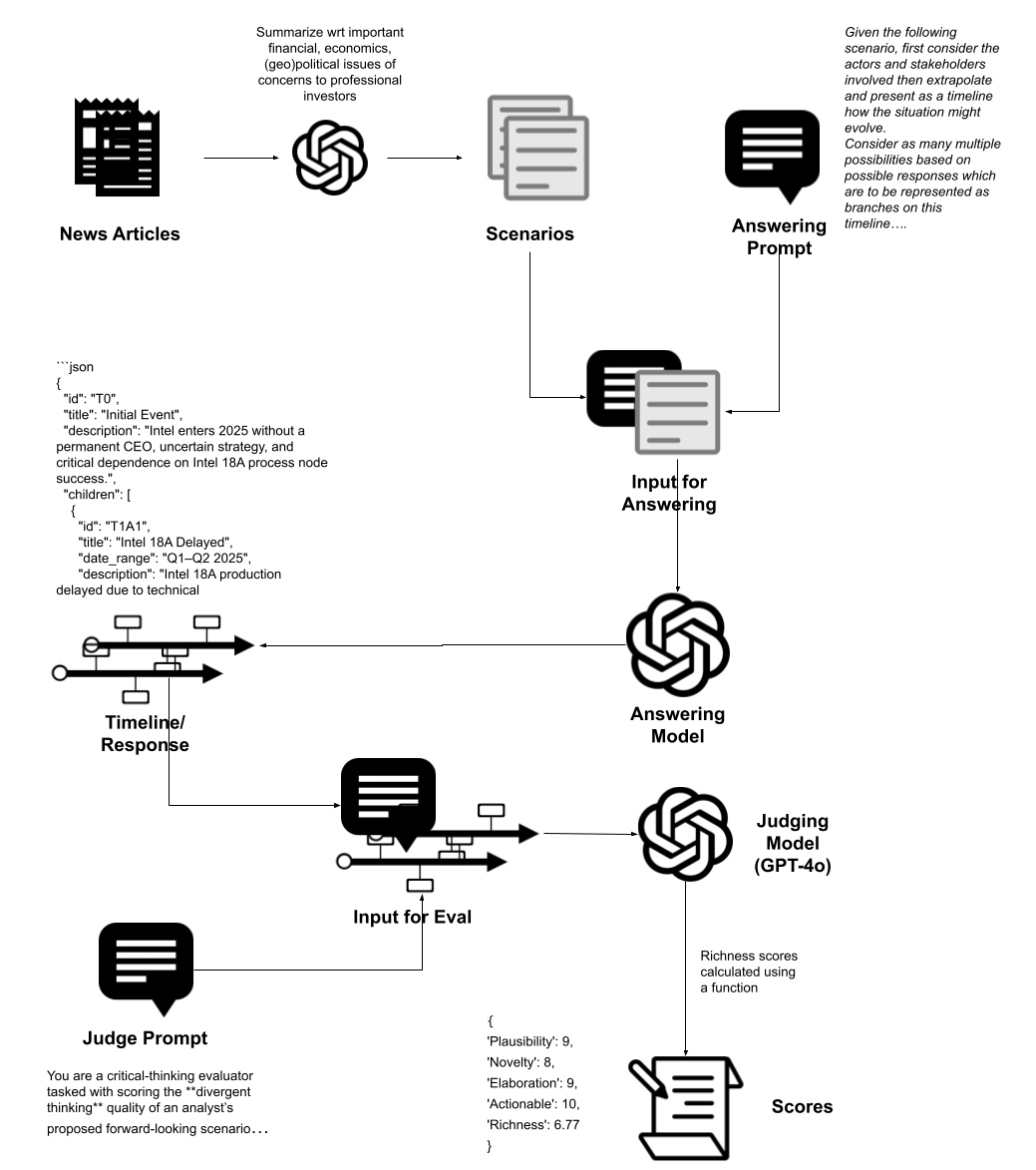} 
    \caption{Pipeline for Dataset Generation for Divergent Thinking Benchmark}
    \label{fig:divergent-thinking-pipeline}
\end{figure}

\noindent The divergent thinking benchmark was created through a structured multi-stage pipeline. The process begins with the curation of real-world financial scenarios from credible sources dated post-training cutoff (after May 2025) to avoid data contamination. Each scenario is a summary of an article with a focus on preserving material financial, economics and political information.

Given a scenario, a single-shot prompt template (see Appendix~\ref{appendix:divergent-input-prompt-template}) is used to create the input prompt into the model to get an output of a branching timeline. This timeline is then combined with a single-shot judging prompt template (see Appendix~\ref{appendix:divergent-judge-prompt-template}) to create an evaluation prompt that utilized GPT-4o to assess Plausibility, Novelty, Elaboration and Actionable. Richness is assessed using a function that treats the timeline as a tree and by calculating graph statistics. This creates a single sample.

A total of 607 distinct scenarios, were evaluated across 14 different models to create a set of 9,380 samples for our analyses below.

\noindent\textbf{Convergent Thinking Dataset}\par\vspace{0.3em}
\begin{figure}
    \centering
    \includegraphics[width=1.0\linewidth]{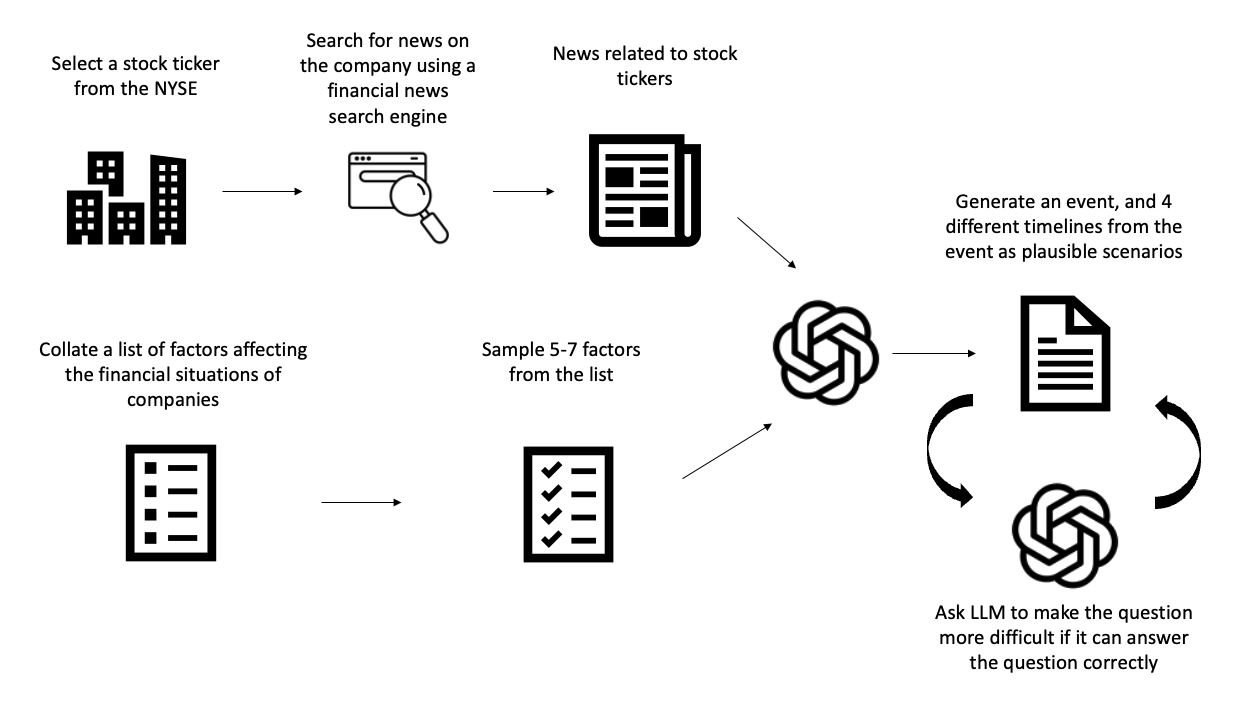}
    \caption{Pipeline for Dataset Generation for Convergent Thinking Benchmark}
    \label{fig:convergent-pipeline}
\end{figure}

\noindent The pipeline for constructing the convergent thinking dataset is shown in Figure \ref{fig:convergent-pipeline} and consists of the following:

\textbf{News search}    Financial news on companies listed on the NYSE are used as the contexts for generating background events for the questions. Using news from 2025 onwards, after the cut-off date for the pre-training data of most of the LLMs, allows us to test their convergent reasoning abilities rather than their memorisation abilities.

\textbf{Factor selection}	 5 random samples are selected from a list of textbook factors which can affect a company’s financial situation. Providing the list of textbook factors ensures that the events and timeline options generated are grounded based on real economic factors, and not baseless situations which neither an LLM nor a human can make sense of.

\textbf{Question generation}    The sampled factors and the search results are given as inputs to GPT-4o using a question generation prompt template (see Appendix~\ref{appendix:convergent-qn-generation-prompt-template}). To make the questions challenging, the model was instructed to use six different adversarial pipelines to generate the event and the questions. The adverserial pipelines are given below:
\begin{itemize}
    \item \textbf{Historical $\beta$-Swap}: Swap one latent driver in a real macro event. Primary effect is to test counterfactual reasoning and causal understanding in complex dynamic systems.
    \item \textbf{Numeric Trip-Wire}: Embed a mini calc (e.g. margin math, FX pass-through). This helps to test quantitative reasoning and precision.
    \item \textbf{Policy Game}: Use a toy policy-model state → next-state logic to spawn one correct seq, one that violates a key equation. This assesses the LLM's ability to understand and apply rule-based systems and logical progression over discrete steps.
    \item \textbf{Cross-Section Confuser}: Include two firms with opposite exposures; only one timeline matches Firm A. This tests the LLM's ability to differentiate and apply general principles to specific, contrasting cases.
    \item  \textbf{Reg-Legal Trap}: Build a legal/reg-process chronology; distractors break one procedural step. This evaluates the LLM's knowledge of domain-specific procedural sequences and regulatory frameworks.
    \item \textbf{Adversarial Self-Play}: Generate many distractors, judge them against the correct path, keep only those within 0.05 plausibility of the correct one. This enhances the difficulty of MCQs by ensuring that distractors are highly confusable and not trivially incorrect. It forces the LLM to make finer distinctions and discourages reliance on simple heuristics.
\end{itemize}

\textbf{Refinement}	The level of difficulty of the questions was increased through a series of refinement steps, taking inspiration from the self-reflection workflow for LLM agents in \cite{renze2024selfreflection}. The LLM was instructed to answer each question. If it was able to answer the question correctly, it would modify the question to make it harder using the same criteria. The prompt for refinement is given in Appendix \ref{appendix:convergent-refinement-prompt-template}. We performed two rounds of refinement to produce our final dataset.

\section{Experiment Setup}
\subsection{Models Evaluated}

\begin{table}[ht]
\centering
\begin{tabularx}{\columnwidth}{|l|X|l|}
\hline
\textbf{Model Family} & \textbf{Azure Model Name} & \textbf{Short Name} \\
\hline
\multirow{2}{*}{Cohere} 
  & cohere-command-a               & cohere\_command\_a \\
  & Cohere-command-r-08-2024       & cohere\_command\_r \\
\hline
\multirow{4}{*}{OpenAI}  
  & gpt-4o                         & gpt4o \\
  & o1                             & o1 \\
  & o1-mini                        & o1\_mini \\
  & o4-mini                        & o4\_mini \\
\hline
\multirow{2}{*}{Mistral} 
  & Mistral-Large-2411             & mistral\_lg \\
  & Mistral-small-2503             & mistral\_sm \\
\hline
\multirow{4}{*}{Llama} 
  & Meta-Llama-3.1-8B-Instruct     & llama3\_8b \\
  & Llama-3.3-70B-Instruct         & llama3\_70b \\
  & Llama-4-Maverick-17B-128E-Instruct-FP8 & llama4\_maverick \\
  & Llama-4-Scout-17B-16E-Instruct & llama4\_scout \\
\hline
\multirow{2}{*}{Others} 
  & DeepSeek-R1                    & deepseek\_r1 \\
  & Phi-4                          & phi4 \\
\hline
\end{tabularx}
\caption{Model Names and Their Shortened Identifiers by Family}
\label{tab:model-names}
\end{table}

We evaluated the list of models shown in \ref{tab:model-names}, which are state-of-the-art at the time of experiment, on the ConDiFi Benchmark, for both divergent and convergent thinking.

\subsection{Divergent Thinking Evaluation}

\noindent\textbf{Evaluation Metrics}\par\vspace{0.3em}

\noindent We performed a multi-dimensional assessment of the timelines/response generated using gpt\_4o via a LLM-As-A-Judge approach that has been widely utilized. Evaluation of the responses is done along five dimensions—Plausibility, Novelty, Elaboration, Actionable, and Richness — selected to capture the multidimensional quality required for such tasks based on our domain knowledge of what matters to professional and high intent retail investors.

These five metrics (refer to Figure~\ref{fig:divergent-thinking-dimensions-result}, and Table~\ref{tab:divergent-thinking-dimensions-result}) collectively balance credibility (Plausibility), originality (Novelty), rigor (Elaboration), utility (Actionable), and structural depth (Richness). We believe they represent a comprehensive and domain-aligned rubric for evaluating scenario-based reasoning in financial contexts, particularly relevant for analysts who must think probabilistically and prepare for fat-tail outcomes.
The judge prompt used for Plausibility, Novelty, Elaboration and Actionable has been designed such that:
\begin{itemize}
    \item A score of $\sim5$ represents the quality of an \emph{average sell-side analyst} note.  
    \item Scores of 9–10 are reserved for the top $5\%$ of timelines exhibiting \emph{exceptional} reasoning or insight.  
    \item A strict penalty scheme deducts points for generic filler phrases, unsupported figures, or substantial repetition.  
    \item Any timeline containing logical impossibilities or temporal inconsistencies is capped at \textbf{3} on \emph{all} dimensions.
\end{itemize}

\textbf{Richness metric.}  
\emph{Richness} is computed automatically by parsing each timeline as a directed tree and mapping graph statistics to a scalar value in $[1,10]$:

\begin{itemize}[leftmargin=2em,labelsep=0.5em]
    \item \textbf{Branching factor}: higher values indicate exploration of multiple plausible futures, akin to how seasoned analysts hedge uncertainty.  
    \item \textbf{Maximum path length}: captures the furthest horizon considered, revealing attention to delayed second- and third-order effects.  
    \item \textbf{Mean path length}: ensures multi-step reasoning is sustained across \emph{most} branches, not only in outliers.  
    \item \textbf{Number of leaf paths}: a large leaf count signals breadth of imagination and explicit modelling of uncertainty.
\end{itemize}

A high \emph{Richness} score therefore denotes \emph{divergent thinking}: breadth of imagination, depth of causal chains, and an uncertainty-aware structure.

\begin{table}[H]
\label{tab:divergent-dimensions}
\centering
\begin{tabular}{lp{5cm}} 
\toprule
\textbf{Metric} & \textbf{Definition} \\
\midrule
Plausibility & Measures whether each node obeys economic cause-effect logic, historical precedent, and macro constraints. \\
Novelty & Captures originality of individual ideas and their interactions, especially second- and third-order effects. \\
Actionable & Assesses the level of detail in each node (e.g., actors, timing, dollar figures) and overall tree structure (e.g., depth and breadth). \\
Elaboration & Evaluates whether the timeline yields specific investment takeaways—e.g., tickers, sectors, asset classes, or hedging triggers. \\
Richness & An automated structural metric based on graph statistics: 1) branching factor, 2) path depth, and 3) scenario breadth—normalized and combined. \\
\bottomrule
\end{tabular}
\caption{Definitions of Evaluation Metrics}
\label{tab:evaluation-metrics}
\end{table}

\noindent The \textit{Richness} metric is computed using four structural components of a scenario tree:

\begin{itemize}[leftmargin=2em]
    \item \textbf{Branching Factor} ($b$): average number of children per non-leaf node, normalized as $s_b = \min\left(\frac{b}{3}, 1.0\right)$
    \item \textbf{Maximum Path Length} ($d_{\max}$): depth of the longest causal chain, normalized as $s_d = \min\left(\frac{d_{\max}}{10}, 1.0\right)$
    \item \textbf{Mean Path Length} ($\bar{d}$): average depth across all leaf paths, normalized as $s_{\bar{d}} = \min\left(\frac{\bar{d}}{7}, 1.0\right)$
    \item \textbf{Breadth} ($p$): total number of leaf paths, normalized on a log scale as $s_p = \min\left(\frac{\log(1 + p)}{\log(101)}, 1.0\right)$
\end{itemize}

The final \textbf{Richness} score is the weighted average of these four components:

\[
\text{Richness} = \min\left(10,\ \max\left(1,\ 10 \times \frac{1}{4}(s_b + s_{\bar{d}} + s_d + s_p)\right)\right)
\]

This formula balances breadth and depth, penalizes shallow or narrow trees, and ensures the score remains in the range \([1, 10]\).

\noindent\textbf{Convergent Thinking Evaluation}\par\vspace{0.3em}

\noindent\textbf{Set-up}
Each model was instructed to answer all questions in the three convergent thinking dataset (original dataset, and refined dataset from each of the three refinement rounds) using the prompt given in Appendix \ref{appendix:convergent-answer-prompt-template}. We used a one-shot chain-of-thought reasoning prompt template where we first tell the model the criteria for good convergent thinking, before giving it an example on how to answer a question, with reasoning.

\noindent\textbf{Metrics}
We define the Convergent Correctness Score (CCS) as the proportion of predictions that exactly match the ground truth over N = 990 evaluation instances. This is computed as:

\[
\text{CCS} = \frac{1}{N} \sum_{i=1}^{N} \left( \text{Pred}_i = \text{GT}_i \right)
\]

Where:
\begin{itemize}
  \item $N = 990$ is the total number of evaluation instances.
  \item $\text{Pred}_i$ is the answer from the LLM for question $i$.
  \item $\text{GT}_i$ is the corresponding ground truth for question $i$.
\end{itemize}

\section{Results and Analysis}
\subsection{Divergent Thinking Analysis}

\noindent\textbf{Overall Results}\par\vspace{0.3em}

\begin{figure}[htbp]
    \centering
    \includegraphics[width=\linewidth]{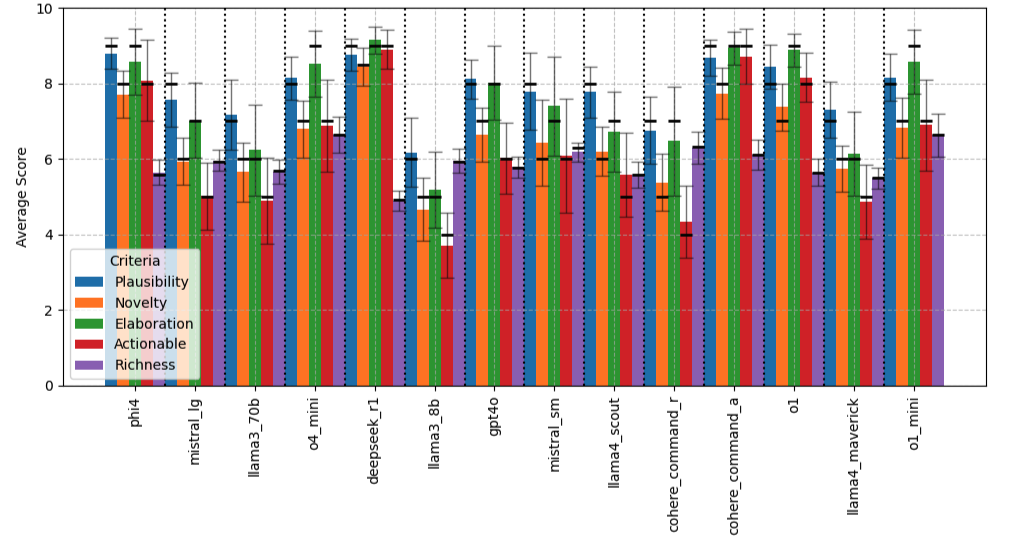} 
    \caption{Average Scores Per Model (Grouped by Model). Black Lines Indicate Median Values}
    \label{fig:divergent-thinking-main-result}
\end{figure}

\begin{table}[t]
\centering
\small
\setlength{\tabcolsep}{2pt} 
\begin{tabular}{lccccccl}
\toprule
\textbf{Model} & \textbf{Plaus.} & \textbf{Novelty} & \textbf{Elab.} & \textbf{Action.} & \textbf{Richness} & \textbf{Overall} \\
\midrule
\texttt{cohere\_command\_a} & 8.67 & 7.74 & 8.94 & 8.72 & 6.11 & 8.04 \\
\texttt{deepseek\_r1}       & 8.76 & 8.45 & 9.15 & 8.90 & 4.89 & 8.03 \\
\texttt{phi4}               & 8.80 & 7.71 & 8.57 & 8.08 & 5.64 & 7.76 \\
\texttt{o1}                 & 8.45 & 7.37 & 8.88 & 8.16 & 5.65 & 7.70 \\
\texttt{o1\_mini}           & 8.16 & 6.82 & 8.56 & 6.90 & 6.63 & 7.41 \\
\texttt{o4\_mini}           & 8.15 & 6.79 & 8.53 & 6.88 & 6.65 & 7.40 \\
\texttt{gpt4o}              & 8.12 & 6.65 & 8.02 & 6.02 & 5.78 & 6.92 \\
\texttt{mistral\_sm}        & 7.79 & 6.43 & 7.41 & 6.08 & 6.18 & 6.78 \\
\texttt{llama4\_scout}      & 7.77 & 6.20 & 6.73 & 5.58 & 5.58 & 6.37 \\
\texttt{mistral\_lg}        & 7.57 & 5.94 & 7.02 & 5.02 & 5.96 & 6.30 \\
\texttt{llama3\_70b}        & 7.17 & 5.66 & 6.24 & 4.89 & 5.67 & 5.93 \\
\texttt{llama4\_maverick}   & 7.31 & 5.73 & 6.14 & 4.87 & 5.50 & 5.91 \\
\texttt{cohere\_command\_r} & 6.75 & 5.38 & 6.48 & 4.34 & 6.30 & 5.85 \\
\texttt{llama3\_8b}         & 6.17 & 4.66 & 5.18 & 3.71 & 5.95 & 5.13 \\
\bottomrule
\end{tabular}
\caption{Model-wise performance across evaluation metrics.}
\label{tab:divergent-thinking-main-result}
\end{table}

\noindent The evaluation results (Figure~\ref{fig:divergent-thinking-main-result}, Table-\ref{tab:divergent-thinking-main-result}) reveal that top-performing models like cohere\_command\_a and deepseek\_r1 consistently outperform others across all divergent thinking dimensions, indicating strong capabilities in generating plausible, novel, elaborate, and actionable scenarios. In contrast, traditional large models like gpt\_4o, mistral\_lg, and llama3\_70b underperform on dimensions like Novelty and Actionability, suggesting limitations in speculative reasoning despite their strength on standard benchmarks. Smaller or older models such as llama3\_8b and cohere\_command\_r struggle across the board. Notably, Richness scores reveal that structural creativity is not solely dependent on model size, highlighting the importance of fine-tuning strategies and prompt alignment in training models for forward-looking, ideation-heavy tasks.

These results raise deeper questions beyond model rankings. Why do models strong in recall struggle with imagining the future? Do they prioritize plausibility over originality, or elaborate without direction? Understanding these trade-offs helps move beyond fluency toward strategic generation. We also ask: can divergent strengths be complementary—paving the way for ensembles that reason more like expert teams than single oracles? We investigate these questions further in the following analysis.

\medskip
\noindent\textbf{Question Difficulty and Variation in Model Performance}\par\vspace{0.3em}

\begin{figure}[htbp]
    \centering
    \includegraphics[width=\linewidth]{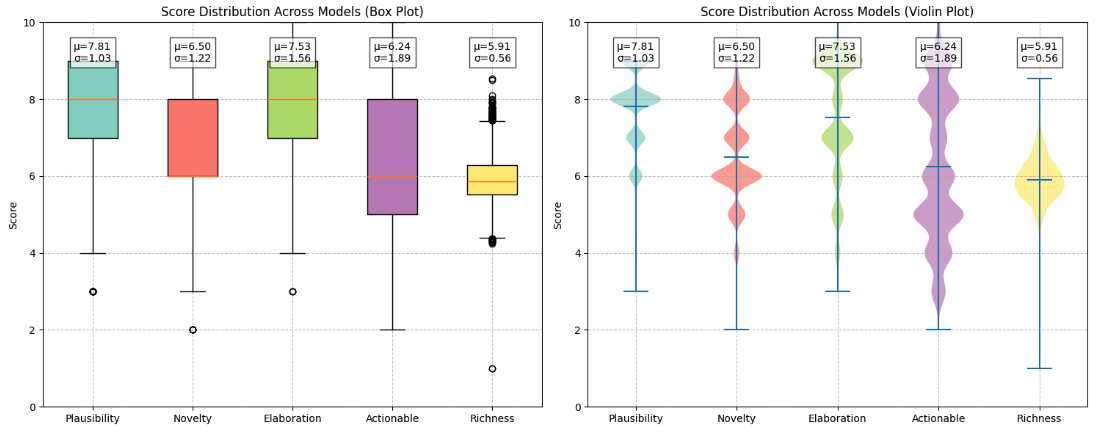} 
    \caption{Scores Distribution of Divergent Thinking Dimensions}
    \label{fig:divergent-thinking-dimensions-result}
\end{figure}

\begin{table}[t]
\small
\centering
\begin{tabular}{lccccc}
\toprule
\textbf{Metric} & \textbf{Mean} & \textbf{Std} & \textbf{Min} & \textbf{Max} & \textbf{Median} \\
\midrule
Plausibility & 7.81 & 1.03 & 3.00 & 9.00 & 8.00  \\
Novelty      & 6.50 & 1.22 & 2.00 & 9.00 & 6.00  \\
Elaboration  & 7.53 & 1.56 & 3.00 & 10.00 & 8.00  \\
Actionable   & 6.24 & 1.89 & 2.00 & 10.00 & 6.00  \\
Richness     & 5.91 & 0.56 & 1.00 & 8.53 & 5.87  \\
\bottomrule
\end{tabular}
\caption{Model score statistics across evaluation dimensions (N = 9,380).}
\label{tab:divergent-thinking-dimensions-result}
\end{table}

\noindent To identify inherently difficult questions in our benchmark, we analyzed score distributions across five dimensions. Low minimums (2–3) and high standard deviations—especially for Actionable ($\sigma$ = 1.89) and Elaboration ($\sigma$ = 1.56)—suggest certain prompts consistently challenge models. Violin plots (Figure~\ref{fig:divergent-thinking-dimensions-result}) show left-skewed distributions with dense low-scoring clusters, reinforcing this. In contrast, Richness shows low variance, likely due to ceiling effects in generative complexity. These patterns indicate the presence of objectively difficult questions, warranting deeper analysis of their semantic or structural traits.

In contrast, Plausibility is bounded by factual coherence, and most models seem quite highly capable of avoiding logically or causally invalid claims. Hence, the tight clustering around the mid-to-high range, with few outliers. 
The lack of multimodal distributions in Richness is because the metric is grounded in structural properties of the timeline tree (e.g., depth, branching, breadth), which most models consistently approximate within a similar range, perhaps due to our experimental settings where models are given a soft budget of 4096 tokens for their response. Future work might look into the impact of higher token budget on the Richness dimension.

We performed a manual examination of some of the easiest and hardest questions and the models’ performance across them to arrive at the following conclusions.

\paragraph{Novelty}
The easiest scenarios (Appendix~\ref{appendix:novelty-examples}) such as CGM or industrial fabric markets—were rich in factual content and structured developments, allowing models to demonstrate lateral thinking through second-order effects (e.g., IP litigation, supply chain ripple effects). Here, most top-tier models (e.g., cohere\_command\_a, deepseek\_r1, phi4) scored consistently high, suggesting they are well-equipped for recombining known economic patterns into original trajectories. However, in more anecdotal or emotionally framed prompts (e.g., personal investing stories), even the best models fell back on clichés or repeated ideas—indicating limited abstraction capabilities when prompts lack structural scaffolding (Appendix~\ref{appendix:novelty-examples}).

\paragraph{Elaboration} 
Elaboration appears to correlate with both model size and instruction tuning quality. \texttt{deepseek\_r1}, \\\texttt{cohere\_command\_a}, and \texttt{o1\_mini} consistently produced well-structured, multi-layered responses with named actors, timeframes, and sector-specific magnitudes. Smaller or weaker instruction-following models (e.g., \texttt{llama3\_8b}, \texttt{llama4\_scout}) often defaulted to surface-level lists or vague generalities. Notably, high elaboration scores were still achievable even in difficult scenarios (Appendix~\ref{appendix:elaboration-examples}) when models had sufficient world knowledge and were trained with strong role-following signals.

\paragraph{Actionable} 
Actionable reasoning surfaced as the most discriminative capability among models. Models like cohere\_command\_a, deepseek\_r1, and phi4 repeatedly translated scenario cues into investable theses or tactical recommendations, especially when tickers or policy levers were mentioned. In contrast, weaker models often failed to extract or operationalize relevant trades—even when the scenarios (Appendix~\ref{appendix:actionable-examples}) contained clear signals.

The results underscore that while prompt quality affects overall scores, model capabilities shape the floor and ceiling of performance, especially on harder prompts. Strong models exhibit robust extrapolation, structured thinking, and speculative abilities, which are critical for financial and strategic applications. Future research should focus on aligning fine-tuning methods with the cognitive demands of divergent, actionable foresight, not just language completion.

\medskip
\noindent\textbf{Intra-model Correlation Patterns: What does each model implicitly prioritize in their divergent thinking?}\par\vspace{0.3em}

\noindent In our evaluation framework for forward-looking scenario timelines, we score each model-generated timeline along five key dimensions: \textbf{Plausibility}, \textbf{Novelty}, \textbf{Elaboration}, \textbf{Actionable}, and \textbf{Richness}. These dimensions reflect not only the logical and creative quality of the scenarios but also their practical utility to professional investors, hedge fund analysts, and equity research analysts.

To better understand how these dimensions interact—and how different models may exhibit strengths or trade-offs—we conduct pairwise correlation analysis across the dimensions. This enables us to:

\begin{itemize}
    \item \textbf{Identify alignment or tension between traits}, e.g., is more \emph{Elaboration} always associated with more \emph{Plausibility}?
    \item \textbf{Detect potential rating biases}, such as whether raters tend to conflate \emph{Richness} with \emph{Elaboration}.
    \item \textbf{Understand trade-offs in model behavior}, helping distinguish models that are creative but impractical from those generating investor-relevant content.
    \item \textbf{Guide model development and tuning} by understanding which dimensions co-vary and which are independent.
\end{itemize}

We focus on correlation pairs with practical alignment (Table~\ref{tab:model-correlations}) to how financial professionals evaluate research—where originality, depth, and rigor are only valuable if they also lead to realistic, actionable strategies.

\begin{table}[t]
\small
\centering
\begin{tabular}{p{2cm}p{2cm}p{3cm}}
\toprule
\textbf{Correlation} & \textbf{Top 3 Models (Descending)} & \textbf{Implication} \\
\midrule
Strong positive correlation between \textbf{Plausibility} and \textbf{Actionable} 
& \texttt{mistral\_sm (0.718)}, \texttt{cohere\_command\_r (0.703)}, \texttt{phi4 (0.683)} 
& Models that generate factually grounded scenarios tend to also produce more investable or decision-relevant insights. This supports the idea that realism is a prerequisite for actionability in investor-oriented timelines. \\
\midrule
High correlation between \textbf{Novelty} and \textbf{Elaboration} 
& \texttt{gpt4o (0.802)}, \texttt{mistral\_sm (0.785)}, \texttt{phi4 (0.783)} 
& Creative scenarios often come with richer, more detailed elaboration. Strong performance here suggests that a model’s lateral thinking is bolstered by its ability to expand ideas through structured narratives. \\
\midrule
Highest correlation between \textbf{Richness} and \textbf{Elaboration} 
& \texttt{o1\_mini (0.237)}, \texttt{mistral\_lg (0.190)}, \texttt{o4\_mini (0.187)} 
& Structural complexity in scenario graphs may co-occur with internal coherence. Strong performance here indicates an ability to simulate diverse, internally coherent futures—an essential trait for strategic foresight and multi-factor reasoning. \\
\bottomrule
\end{tabular}
\caption{Top model-level correlations between evaluation dimensions and their implications.}
\label{tab:model-correlations}
\end{table}

Note that should not be interpreted as an indicator of each model’s performance in the highlighted dimensions, but rather its tendency or tension to align outputs along certain dimensions. 

\medskip
Overall, the analysis reveals clear patterns in how models balance traits essential for generating high-quality financial scenarios. Strong Plausibility–Actionable alignment shows realism underpins actionable insight. High Novelty–Elaboration correlation highlights that creativity often comes with rich narrative development. Yet the weak Richness–Elaboration correlation, even among top models, exposes a gap: models can elaborate fluently without producing structurally rich, branching futures. Closing this gap is vital for enabling LLMs to simulate complex, multi-factor scenarios—especially in finance, where anticipating interdependent outcomes is crucial.

Refer to the Appendix (\ref{appendix: correlation-across-dimensions}) for full set of correlation matrices.

\medskip
\noindent\textbf{Inter-model Distance: Which Models Behave Similarly, Which Ones Are Complementary?}\par\vspace{0.3em}

\noindent In this study, we conducted an inter-model distance analysis using the Frobenius norm of correlation matrices across five evaluation dimensions: \textbf{Plausibility}, \textbf{Novelty}, \textbf{Elaboration}, \textbf{Actionable}, and \textbf{Richness}. Each model’s correlation matrix captures how these dimensions co-vary in its output—essentially offering a behavioral fingerprint of the model’s reasoning tendencies.

\medskip

The Frobenius norm provides an intuitive and stable measure of overall difference between two matrices, taking into account all pairwise interactions between dimensions. Unlike element-wise comparisons, it allows for a compact and holistic representation of structural similarity.

Large distances indicate dissimilar “reasoning fingerprints”; small distances suggest overlapping behavior. While the metric does not itself reveal which model is ‘better’, it helps us
spot (i) clusters of models that approach the task in similar ways and (ii) outliers whose unique correlation patterns may provide complementary strengths in an ensemble.

\begin{figure}[htbp]
    \centering
    \includegraphics[width=\linewidth]{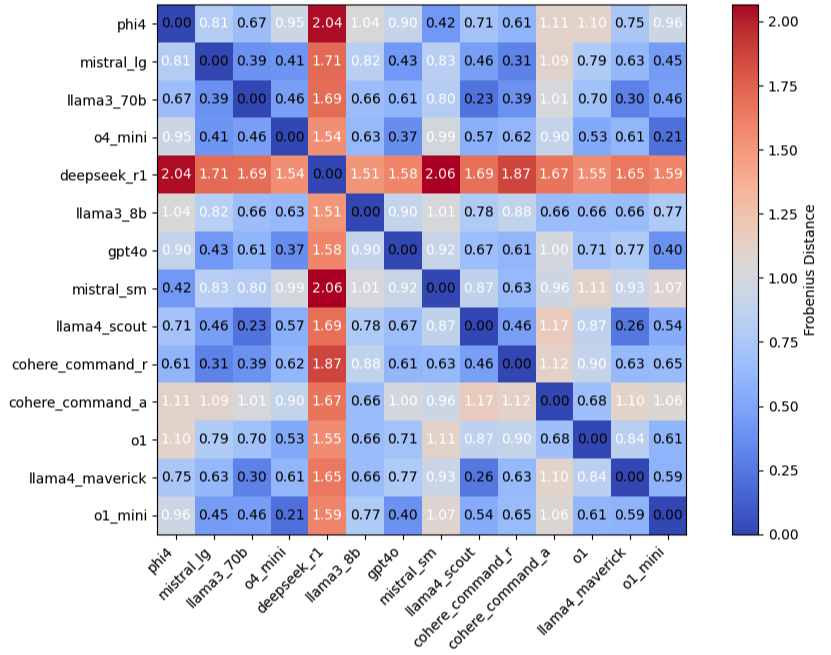} 
    \caption{Frobenius Norm of Inter-Model Correlations}
    \label{fig:frobeniums-heatmap}
\end{figure}

\begin{table*}[t]
\centering
\caption{Frobenius Norm Distance Between Model Correlation Matrices (Split View)}
\label{tab:frobenius-aligned}
\scriptsize
\begin{subtable}[t]{0.49\textwidth}
\centering
\caption{Part 1: Models 1–7}
\begin{tabular}{lrrrrrrr}
\toprule
& phi4 & mistral\_lg & llama3\_70b & o4\_mini & deepseek & llama3\_8b & gpt4o \\
\midrule
phi4 & 0.00 & 0.81 & 0.67 & 0.95 & 2.04 & 1.04 & 0.90 \\
mistral\_lg &       & 0.00 & 0.39 & 0.41 & 1.71 & 0.82 & 0.43 \\
llama3\_70b &       &      & 0.00 & 0.46 & 1.69 & 0.66 & 0.61 \\
o4\_mini &          &      &      & 0.00 & 1.54 & 0.63 & 0.37 \\
deepseek &          &      &      &      & 0.00 & 1.51 & 1.58 \\
llama3\_8b &         &      &      &      &      & 0.00 & 0.90 \\
gpt4o &             &      &      &      &      &      & 0.00 \\
\bottomrule
\end{tabular}
\end{subtable}
\hfill
\begin{subtable}[t]{0.49\textwidth}
\centering
\caption{Part 2: Models 8–14}
\begin{tabular}{lrrrrrrr}
\toprule
& mistral\_sm & scout & cmd\_r & cmd\_a & o1 & maverick & o1\_mini \\
\midrule
mistral\_sm & 0.00 & 0.87 & 0.63 & 0.96 & 1.11 & 0.93 & 1.07 \\
scout       &      & 0.00 & 0.46 & 1.17 & 0.87 & 0.26 & 0.54 \\
cmd\_r      &      &      & 0.00 & 1.12 & 0.90 & 0.63 & 0.65 \\
cmd\_a      &      &      &      & 0.00 & 0.68 & 1.10 & 1.06 \\
o1          &      &      &      &      & 0.00 & 0.84 & 0.61 \\
maverick    &      &      &      &      &      & 0.00 & 0.59 \\
o1\_mini    &      &      &      &      &      &      & 0.00 \\
\bottomrule
\end{tabular}
\end{subtable}
\end{table*}

Notably, deepseek\_r1 emerges as a distinct outlier, exhibiting the highest average Frobenius distance from other models (Figure~\ref{fig:frobeniums-heatmap}). This suggests a unique internal correlation structure, likely stemming from its unconventional training approach that emphasizes reinforcement learning without reliance on labeled data. By prioritizing correct outputs over human-like reasoning, deepseek\_r1 may develop internal representations that differ significantly from models trained with supervised fine-tuning or reinforcement learning with human feedback.

Conversely, models within the LLaMA family, such as llama3\_70b, llama4\_maverick, and llama4\_scout, display low Frobenius distances among themselves, indicating consistent internal structures. This consistency can be attributed to shared architectural designs and training methodologies, including the use of Grouped-Query Attention and instruction tuning with supervised fine-tuning and reinforcement learning with human feedback.

Cohere’s Command models, particularly cohere\_command\_a and cohere\_command\_r, exhibit notable distances from other models, reflecting their unique training focuses. cohere\_command\_a, for instance, is optimized for tool use, retrieval-augmented generation, and multilingual tasks, which may influence its internal evaluation metrics differently compared to models with general-purpose training objectives.

Understanding these distances aids in selecting models with complementary evaluation behaviors for ensemble methods, potentially enhancing overall performance. Models with unique training approaches, like deepseek\_r1, may offer novel perspectives but could also introduce alignment challenges due to their divergent internal structures. Recognizing the training focuses of models allows for better alignment between model capabilities and specific application requirements.

\begin{table}[t]
\centering
\caption{Similar Model Pairs}
\label{tab:similar-models}
\small
\begin{tabular}{p{2cm}p{1cm}p{4cm}}
\toprule
\textbf{Model Pairs} & \textbf{Distance} & \textbf{Interpretation} \\
\midrule
o4\_mini, o1\_mini & 0.214 & Very close correlation behavior; not surprising given same base family or similar tuning. \\
llama3\_70b, llama4\_maverick & 0.302 & Very similar patterns of co-variation across dimensions—despite generational name difference. \\
llama3\_70b, llama4\_scout & 0.230 & LLaMA variants show structural consistency in evaluation metrics. \\
phi4, mistral\_sm & 0.424 & Despite different branding, shows strong similarity—may reflect overlapping training objectives. \\
\bottomrule
\end{tabular}
\end{table}

\begin{table}[t]
\centering
\caption{Complementary Model Pairs}
\label{tab:complementary-models}
\small
\begin{tabular}{p{2cm}p{1cm}p{4cm}}
\toprule
\textbf{Model Pairs} & \textbf{Distance} & \textbf{Interpretation} \\
\midrule
deepseek\_r1, mistral\_sm & 2.06 & Highly divergent—DeepSeek’s internal dimension correlations behave very differently. \\
deepseek\_r1, phi4 & 0.302 & Shares very similar co-variation structure—despite architectural and training differences. \\
deepseek\_r1, cohere\_command\_a & 0.230 & Surprisingly similar structure, suggesting possible complementary modeling choices. \\
\bottomrule
\end{tabular}
\end{table}

\begin{figure}[htbp]
    \centering
    \includegraphics[width=\linewidth]{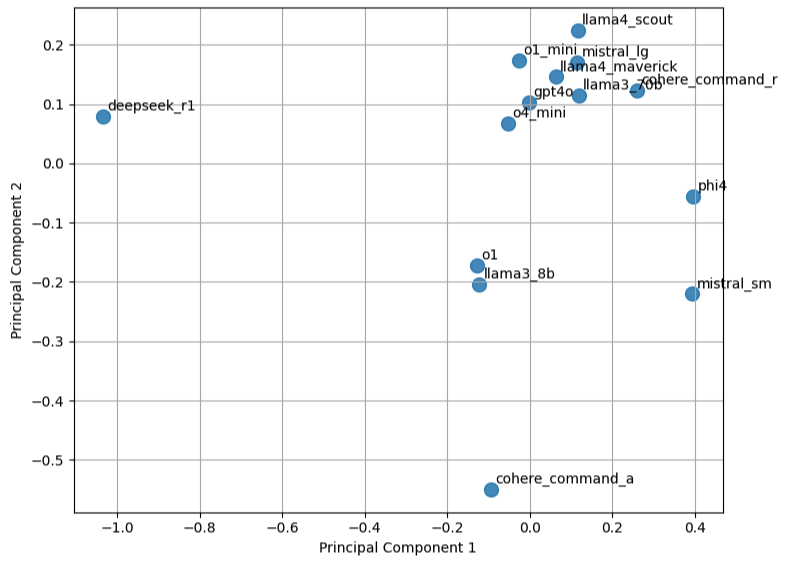} 
    \caption{PCA of Model Correlation Structures}
    \label{fig:pca-correlation}
\end{figure}

To better understand the global structure of model behavior across evaluation dimensions, we conducted a Principal Component Analysis (PCA) on the full set of model scores. PCA serves as a dimensionality reduction technique that transforms the original five evaluation metrics—Plausibility, Novelty, Elaboration, Actionable, and Richness—into a set of orthogonal components, capturing the directions of greatest variance in model performance (Figure~\ref{fig:pca-correlation}).

Principal Component 1 (PC1): Alignment with Human Preference
Notably, we observe that deepseek\_r1 is an outlier from the rest along this dimension. By looking at the cluster of high models in the top right corner and based on the known portions of their training methodologies, we believe this possibly represents some form of alignment with human preference or the pre-training methodology. deepseek\_r1 is famously known to have been trained with direct reinforcement learning as opposed to RLHF of some variety that relies on human preference data, likely explaining its position as an outlier here.

Principal Component 2 (PC2): Reasoning Style
We see more than two distinct groupings of models here along PC2. With phi4, mistral\_sm, o1 and llama3\_8b scoring in the middle, and 
\\cohere\_command\_a scoring at the bottom as an extreme outlier. The other models, including deepseek\_r1 have very similar scores. Our best guess is that PC2 represents some kind of reasoning style. phi4 is known to have utilized a significant amount of synthetic data in both its pre-training and fine-tuning phase. Whereas the o1 model is also known to have utilized synthetic data in the fine-tuning phase. The synthetic data generation process is likely to have influenced their reasoning styles versus other models that make little to no use of such processes. The cohere\_command\_a is known to have an especially elaborate fine-tuning/post-training strategy with multi-stage reinforcement learning and synthetic data amongst others, which might explain its especially great deviation from the other models.  

These analyses reveal how models differ in reasoning style, not just performance, informing model selection for tasks like financial analysis or strategic foresight that demand divergent thinking. Notably, DeepSeek-R1’s distinct behavior along the first principal component likely reflects its training on reasoning-focused synthetic data and RLHF for chain-of-thought tasks—suggesting that fine-tuning can shape a model’s cognitive style. This opens avenues for designing training regimens that cultivate specific reasoning traits for creative or strategic applications.

\subsection{Convergent Thinking Analysis}
\noindent\textbf{Overall Results}\par\vspace{0.3em}
\begin{table}[t]
\begin{tabular}{lrrr}
\toprule
\multicolumn{1}{l}{\textbf{Model}}   & \multicolumn{1}{l}{\textbf{Original}} & \multicolumn{1}{l}{\textbf{Refinement 1}} & \multicolumn{1}{l}{\textbf{Refinement 2}} \\
\midrule
llama4\_maverick & 0.9547                                & 0.9234                                    & 0.7232                                    \\
o1                                     & 0.9245                                & 0.8962                                    & 0.7141                                    \\
llama4\_scout         & 0.9567                                & 0.9375                                    & 0.7101                                    \\
llama3\_70b                 & 0.9406                                & 0.9254                                    & 0.701                                     \\
deepseek\_r1                            & 0.9355                                & 0.8931                                    & 0.701                                     \\
gpt\_4o                                 & 0.9376                                & 0.9022                                    & 0.698                                     \\
phi\_4                                  & 0.9204                                & 0.8952                                    & 0.6949                                    \\
o4\_mini                                & 0.8704                                & 0.8579                                    & 0.6898                                    \\
cohere\_command\_a                       & 0.8983                                & 0.8659                                    & 0.6815                                    \\
o1\_mini                                & 0.872                                 & 0.8276                                    & 0.6737                                    \\
mistral\_sm                     & 0.856                                 & 0.8468                                    & 0.6636                                    \\
mistal\_lg                     & 0.8648                                & 0.8325                                    & 0.6431                                    \\
llama3\_8b             & 0.5389                                & 0.5626                                    & 0.4752                                    \\
cohere\_command\_r               & 0.4931                                & 0.4969                                    & 0.4161                                    \\
\textbf{Average}                       & \textbf{0.8545}                       & \textbf{0.8331}                           & \textbf{0.6561}                          
\end{tabular}
\caption{Models' CCS Scores for Convergent Datasets}
\label{tab:convergent-overall-results}
\end{table}



\noindent Table \ref{tab:convergent-overall-results}, sorted in descending order of CCS for \textit{Refinement 2}, shows the convergent thinking ability of the different models. One key observation is that the refinements made the questions more difficult. After the first round of refinement, there was a slight fall in the CCS from an average score of 85.5\% to 83.3\% (2.57\% decrease). After the second round, the difficulty increased significantly and the average CCS score further fell from 83.3\% to 65.6\% (21.2\% decrease). Another key observation is that the benchmark (refinement 2) is sufficiently difficult for modern LLMs with an upper bound score of 72.32\% and a lower bound score of 41.61\% for the LLMs evaluated.

In terms of individual model performance, the Llama series models perform exceptionally well, with 3 out of the 4 models evaluated taking the first, second, and fourth spots, the only exception being llama3\_80b which is much older and smaller than the others. OpenAI models also performed well with scores ranging from 67.4\% to 71.4\%, with the best performing model among them, o1, ranking second. The larger models, o1 and gpt\_4o, scored better than the mini models. It can be seen that a “full-sized” reasoning model (o1) performed better than a non-reasoning model (gpt\_4o), but mini models, despite being optimised for reasoning, did not perform better than a “full-sized” non-reasoning model (gpt\_4o). Another model optimised for reasoning, deepseek\_r1, ranked high in joint fourth spot with a score of 70.1\%.

\noindent\textbf{Question Answerability}\par\vspace{0.3em}
\begin{figure}
    \centering
    \includegraphics[width=1\linewidth]{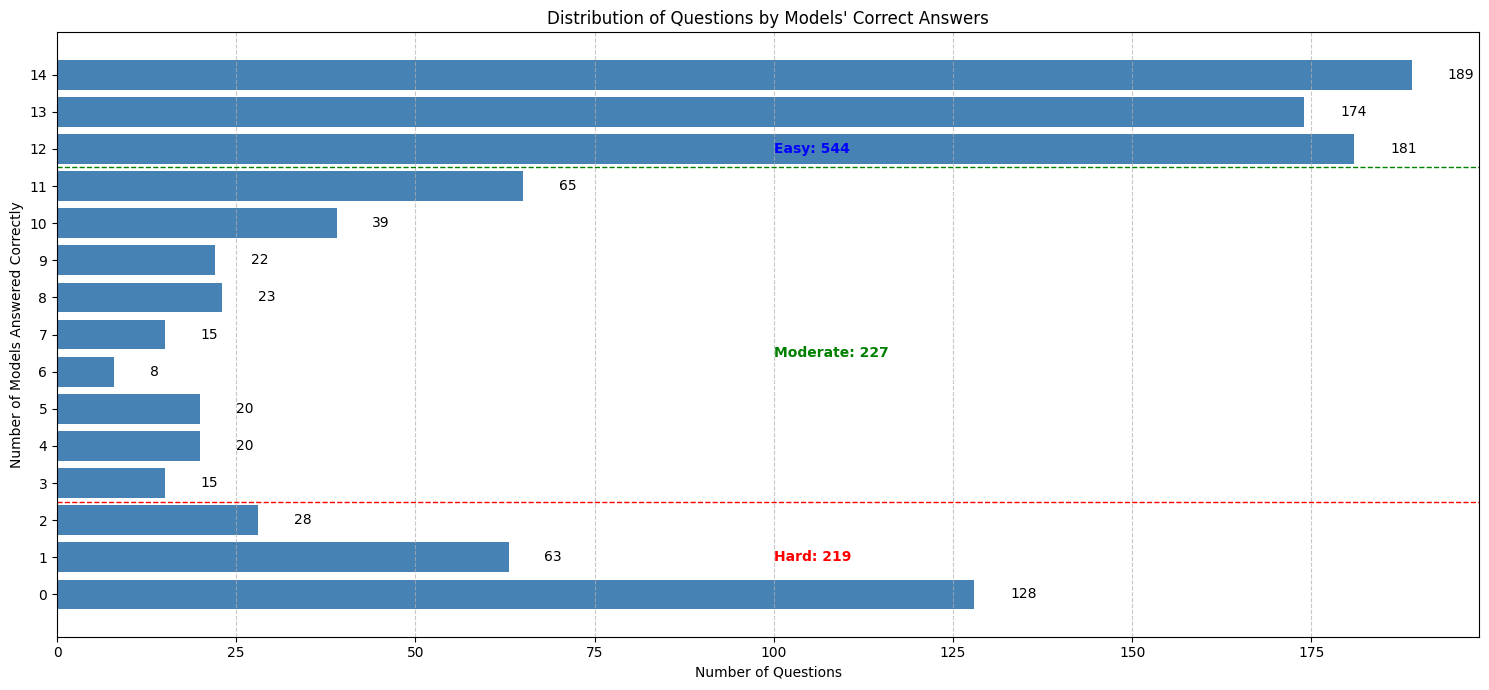}
    \caption{Breakdown of Question Answerability by the different models}
    \label{fig:question-answerability}
\end{figure}

\noindent Figure \ref{fig:question-answerability} shows the breakdown of the difficulty of the questions. At the extreme ends, 128 questions could not be correctly answered by any of the models and 189 questions were correctly answered by all the models. Categorizing them into Hard, Moderate, and Easy, we have a tiered breakdown whereby a model with poor convergent thinking ability will score 55\% or less on the benchmark (can only answer the easy questions), one with good ability will score between 55\% and 78\% (can answer all the easy questions and some moderate questions), and one with excellent ability will score above 78\% (can answer all the easy, moderate questions, and some hard questions).

\noindent\textbf{Error Analysis}\par\vspace{0.3em}

\noindent To take a deeper look at the difficulty of the questions, we analysed the errors of llama4\_maverick, the current best-performing model on the convergent thinking dataset. We used gpt\_4o as an assessor to explain the error using the prompt in \ref{appendix:convergent-error-analysis-prompt}. The summaries of the errors are shown in Appendix \ref{appendix:convergent-thinking errors}. Using the error analyses, we grouped them into six categories as follows and show the composition of each error (percentages do not add up to 100\% as each error can belong to multiple categories):
\begin{itemize}
    \item \textbf{Misinterpretation of Scenario Nuances and Key Information (68\%)}  The model failed to grasp the core elements, specific details (e.g., "procedural delays," "compounded challenges," "subsidy uncertainty"), or the intended focus/emphasis of the scenario. This includes misinterpreting specific dynamics like competitive, regulatory, or financial aspects.

    \item \textbf{Incorrect Prioritization or Weighting of Factors (62\%)} The model incorrectly weighed the importance of different factors presented. This often involves overemphasizing certain aspects (e.g., operational balance, temporal coherence, positive milestones, short-term benefits) while neglecting or underemphasizing others (e.g., regulatory penalties, investor sentiment, strategic partnerships, long-term goals).

    \item \textbf{Overlooking Specific Critical Details (28\%)}  The model missed a crucial, explicitly stated piece of information or detail within the scenario or timeline options that was key to the correct answer.
    
    \item \textbf{Bias Towards Optimism / Positive Outcomes (18\%)} The model showed a tendency to select timelines with more positive, favorable, or straightforward outcomes, even when the scenario emphasized challenges, negative signals, ambiguity, or a more cautious outlook.

    \item \textbf{Flawed Logical or Causal Reasoning (14\%)}  The model made errors in understanding the sequence of events, cause-and-effect relationships, or logical entailment between different parts of the scenario or timeline options. This includes misinterpreting causal links or the logical progression of events.

    \item \textbf{Misunderstanding of Task Objective or Evaluation Criteria (7\%)}  The model seemed to misunderstand the underlying goal of the evaluation or the specific criteria it was supposed to be judged against. This includes focusing on "realism" when the test intended to evaluate entailment under unrealistic assumptions, or not recognizing the test's intent to highlight specific types of challenges or pivots.
\end{itemize}

\section{Limitations}

\begin{enumerate}[leftmargin=1.5em]
    \item \textbf{Domain specificity} – We restrict scenarios to post-May 2025 U.S. and major-market equities to minimize training-data leakage. However, syndicated news and reference drift make full isolation from pretraining data unlikely.
    
    \item \textbf{LLM-as-judge constraints} – Divergent-thinking scores are generated by GPT-4o. Despite penalty catalogues and bias checks, such automated evaluation inherently inherits the model's own biases and reasoning artifacts. Human audits remain essential for robustness.

    \item \textbf{Richness metric bias} – Our graph-based \textit{Richness} metric captures structural creativity ($\rho \approx 0.56$ with human ratings), but may reward breadth over depth, inflating scores for wide but shallow trees.

    \item \textbf{Unobservable training overlap} – Proprietary models may have ingested similar data via undisclosed refresh cycles, despite scenarios being post-cutoff.

    \item \textbf{Limited model scope} – We evaluate 14 public LLMs across four families. Closed-source or domain-fine-tuned models—especially Claude and Gemini—are omitted due to access constraints.

    \item \textbf{Token ceiling} – Outputs were capped at 4096 tokens. Early evidence suggests that longer generations may boost \textit{Richness}, especially for GPT-4-class models.

    \item \textbf{Prompting minimalism} – For consistency, we used single-shot prompts without tool use or few-shot examples. Richer prompting could unlock deeper reasoning capabilities.

    \item \textbf{U.S.-centric framing} – Actionability assumes U.S. financial norms and may not generalize to other regulatory environments or asset classes.

    \item \textbf{Decoding fixed} – We did not explore temperature variation or nucleus sampling. These can affect creativity, determinism, and score variance.
\end{enumerate}

\section{Conclusion}

We introduce \textbf{ConDiFi}, the first benchmark to evaluate both divergent and convergent reasoning in LLMs within a real-world financial context. While standard evaluations focus on factual or procedural accuracy, ConDiFi emphasizes reasoning styles aligned with high-stakes decision-making—where analysts must speculate, plan, and elaborate under uncertainty.

Spanning 607 macro-financial prompts and 990 multi-hop MCQs, ConDiFi reveals asymmetries in model behavior: fluency and plausibility do not necessarily imply novelty or strategic utility. Some models elaborate convincingly but fail to imagine meaningful futures.

Our key contribution lies in the structural evaluation of divergent thinking. We model scenario outputs as branching timelines and score them using graph-based \textit{Richness}—a novel metric capturing creative structure. Correlation analyses and Frobenius clustering further illuminate reasoning patterns beyond raw scores.

\textbf{Future directions} include releasing the full dataset and scoring scripts, integrating human evaluations and causal tracing, extending to global markets, and probing model behavior under varied prompting and decoding strategies.

ConDiFi offers a path toward more cognitively grounded, domain-sensitive evaluation—helping LLMs evolve from fluent explainers to strategic reasoners in domains where the cost of error is high.

\section*{Disclaimer} 
AI tools were used for post-editing and grammatical corrections when writing this paper. The authors are solely responsible for the final content.


\appendix









\appendix
\onecolumn
\section{Appendix I - Divergent Thinking System and User Prompts}

\subsection*{Divergent Thinking System Prompt}
\begin{tcolorbox}[colback=gray!5!white,colframe=black!75!white,title=System Prompt: Investment Analyst Persona]
You are a 300 IQ investment analyst with extensive academic and professional background in economics, financial markets, and politics. Your role is to help your portfolio manager and team with deep fundamental analysis as both generalists and specialists across sectors and asset classes to produce alpha and monitor portfolio risk.

Your job requires that you think logically but also originally and creatively to spot hidden risks and opportunities that other experts may have overlooked. Examples include:

\textbf{Stanley Druckenmiller – Shorting Tech Bubble}
\begin{itemize}
  \item \textbf{Bet}: Soros Fund shorted the dot-com bubble before the 2000 crash.
  \item \textbf{Rationale}: Recognized extreme valuations and speculative mania despite missing out on late-stage upside.
  \item \textbf{Outcome}: Massive profits post-bubble burst.
  \item \textbf{Why intellectual}: Showed discipline, valuation sense, and courage to go against market euphoria.
\end{itemize}

\textbf{John Paulson – Subprime Meltdown}
\begin{itemize}
  \item \textbf{Bet}: Built synthetic CDOs to short subprime mortgage exposure.
  \item \textbf{Rationale}: Saw the unsustainable structure of MBS markets and mispricing of risk.
  \item \textbf{Outcome}: \textasciitilde\$15B in gains.
  \item \textbf{Why intellectual}: Engineered financial instruments to express his view, ahead of the curve.
\end{itemize}

\textbf{David Tepper – Buying Bank Stocks}
\begin{itemize}
  \item \textbf{Bet}: Bought distressed financial stocks like Citi and Bank of America when most investors were fleeing.
  \item \textbf{Rationale}: Believed Fed liquidity and TARP would backstop the system.
  \item \textbf{Outcome}: Appaloosa Management made billions in recovery.
  \item \textbf{Why intellectual}: Contrarian thinking, macro signal synthesis, and balance sheet analysis.
\end{itemize}
\end{tcolorbox}

\onecolumn
\subsection*{Divergent Thinking Prompt}
\label{appendix:divergent-input-prompt-template}
\begin{tcolorbox}[
  colback=gray!5!white,
  colframe=black!75!white,
  title=Answering Prompt: Timeline Generation,
  fonttitle=\bfseries,
  coltitle=white,
  arc=2mm,
  boxrule=0.4pt,
  breakable  
]
Given the following scenario, first identify the actors and stakeholders involved. Then extrapolate how the situation may evolve, expressed as a timeline. Represent multiple plausible responses as branching paths.

Ensure the timeline is of \textbf{HIGH QUALITY}, using the following criteria:

\begin{itemize}
  \item \textbf{Plausibility} — Is the timeline economically, politically, and financially coherent?
  \begin{itemize}
    \item High: Paths are backed by macro data, expert knowledge, and historical precedent (e.g., “monetary tightening leads to recession, triggering political backlash”)
    \item Low: Logical violations or unrealistic events (e.g., hyperinflation leads to rate cuts)
  \end{itemize}

  \item \textbf{Novelty} — Are the scenarios creatively reasoned and non-obvious?
  \begin{itemize}
    \item High: Creative, counterintuitive insights, rare risks, or novel feedback loops (e.g., “trade war accelerates semiconductor realignment via Middle East sovereign funds”)
    \item Low: Follows headlines or boilerplate analysis with no original paths
  \end{itemize}

  \item \textbf{Elaboration} — Are the nodes specific and refined?
  \begin{itemize}
    \item High: Each node acts as a mini case study with actors, numbers, feedbacks, or sector detail (e.g., “Volkswagen reallocates \$1.2B R\&D from EVs to hybrids”)
    \item Low: Generic (e.g., “market reacts”) or lacks causal clarity
  \end{itemize}

  \item \textbf{Actionable} — Can the timeline inform tradeable decisions?
  \begin{itemize}
    \item High: Includes sectors, tickers, or entry/exit triggers (e.g., “short EU auto suppliers”, “long US rail freight”)
    \item Low: Insightful but lacks investment utility
  \end{itemize}
\end{itemize}

\textbf{Format:} Use ASCII to represent the timeline tree in JSON.

\medskip
\textbf{Example Scenario:} \\
WASHINGTON/DETROIT, April 29 (Reuters) – U.S. President Donald Trump signed two orders to soften auto tariff impacts with credits and material levy relief...

\medskip
\textbf{Example Output:}
\begin{flushleft}
\scriptsize
\begin{verbatim}
{
  "id": "T0",
  "title": "Initial Event",
  "description": "German inflation data released at 2.2%",
  "children": [
    {
      "id": "T1A1",
      "title": "ECB maintains dovish stance",
      "date_range": "May",
      "description": "ECB maintains dovish stance, keeps rates low",
      "children": [
        {
          "id": "T1A1.1",
          "title": "German banks increase lending",
          "description": "German banks benefit from low rates, increase lending"
        },
        {
          "id": "T1A1.2",
          "title": "Euro weakens",
          "description": "Euro weakens, boosting exports"
        },
        {
          "id": "T1A1.3",
          "title": "Moderate industrial growth",
          "description": "German industry reports moderate growth"
        },
        {
          "id": "T1A1.4",
          "title": "Bundesbank caution",
          "description": "Bundesbank signals caution on inflation",
          "children": [
            {
              "id": "T2A1",
              "title": "Industry adapts to inflation",
              "date_range": "June–August",
              "description": "Companies pass on higher costs; unions request wage hikes; consumer shifts emerge",
              "children": [
                {
                  "id": "T3A1",
                  "title": "Markets stabilize",
                  "date_range": "Q3–Q4",
                  "description": "Eurozone growth outlook improves; German stocks outperform",
                  "children": [
                    {
                      "id": "T4A1",
                      "title": "Global growth stabilizes",
                      "date_range": "Q4–2025",
                      "description": "Global trade rebounds; emerging markets benefit",
                      "children": [
                        {
                          "id": "T5A1",
                          "title": "New normal",
                          "date_range": "2025+",
                          "description": "Stable growth across EU/US; ECB remains dovish"
                        }
                      ]
                    }
                  ]
                }
              ]
            }
          ]
        }
      ]
    },
    {
      "id": "T1A2",
      "title": "ECB tightens monetary policy",
      "date_range": "May",
      "description": "ECB tightens policy in response to inflation",
      "children": [
        {
          "id": "T1A2.1",
          "title": "Euro strengthens",
          "description": "Euro strengthens, hurting exports"
        },
        {
          "id": "T1A2.2",
          "title": "Industrial slowdown",
          "description": "German industry reports slower growth"
        },
        {
          "id": "T1A2.3",
          "title": "Bundesbank warns",
          "description": "Bundesbank warns of inflation risks"
        },
        {
          "id": "T1A2.4",
          "title": "Consumers cut back",
          "description": "German consumers cut back on spending",
          "children": [
            {
              "id": "T2A2",
              "title": "Industry opposes ECB policy",
              "date_range": "June–August",
              "description": "Job cuts, political backlash, delayed consumption",
              "children": [
                {
                  "id": "T3A2",
                  "title": "Markets unstable",
                  "date_range": "Q3–Q4",
                  "description": "Growth outlook declines; ECB remains hawkish",
                  "children": [
                    {
                      "id": "T4A2",
                      "title": "Global recession risk rises",
                      "date_range": "Q4–2025",
                      "description": "Trade slows; emerging markets struggle",
                      "children": [
                        
                          "id": "T5A2",
                          "title": "Global downturn",
                          "date_range": "2025+",
                          "description": "Recession conditions persist; ECB maintains rate hikes"
                        }
                      ]
                    }
                  ]
                }
              ]
            }
          ]
        }
      ]
    }
  ]
}
\end{verbatim}
\end{flushleft}

\vspace{0.5em}
\textbf{[Given Scenario]} \{\} \\
\textbf{[Output in JSON]} \{\}
\end{tcolorbox}

\subsection*{System Prompt: Investment Judge Persona}
\begin{tcolorbox}[
  colback=gray!5!white,
  colframe=black!75!white,
  title=System Prompt: Portfolio Manager Evaluator,
  fonttitle=\bfseries,
  coltitle=white,
  arc=2mm,
  boxrule=0.4pt,
  breakable
]
You are a 300 IQ top-performing portfolio manager who thinks critically about the investment ideas presented to you in order to arrive at
informed decisions and actions. You are strict with any research and investment proposals presented to you, focusing on fine details and 
easily spotting logical inconsistencies, weak assumptions, or unsupported claims.
\end{tcolorbox}

\subsection*{Judge Prompt: Scoring Divergent Thinking}
\label{appendix:divergent-judge-prompt-template}
\begin{tcolorbox}[
  colback=gray!5!white,
  colframe=black!75!white,
  title=User Prompt: Divergent Thinking Evaluation,
  fonttitle=\bfseries,
  coltitle=white,
  arc=2mm,
  boxrule=0.4pt,
  breakable
]
You are a critical-thinking evaluator tasked with scoring the \textbf{divergent thinking} quality of an analyst’s proposed forward-looking scenario.

The analyst has provided a \textbf{branching timeline}, mapping possible ways a macroeconomic or geopolitical scenario might evolve. Each branch represents a plausible world trajectory.

\bigskip
\textbf{1. Scoring Framework}
\begin{itemize}
    \item \textbf{5 ± 0.5} = average sell-side note  
    \item Reserve \textbf{9–10} for the top \textbf{5\%} of timelines you have ever seen  
    \item Reserve \textbf{1–2} for timelines that would mislead an investor
\end{itemize}

Rate \textbf{strictly} and apply the penalties below even if only one branch is affected.

\bigskip
\textbf{2. Penalty Catalogue}
\begin{itemize}
  \item \textbf{Generic filler node} (e.g., "market reacts", "tensions rise", "empty buzzwords"): \textbf{-1} per 3 occurrences
  \item \textbf{Unsupported or impossible macro figure/date}: \textbf{-2} each
  \item \textbf{Repeating more than 30\% of baseline timeline ideas}: \textbf{-2} total
\end{itemize}

\bigskip
\textbf{3. Dimension Rubric (1–10)}
\begin{itemize}
    \item \textbf{Plausibility} — 10 = macro-logical and historically grounded; \\
    \phantom{--} 5 = plausible but thin; <3 = breaks cause-effect or factual reality  
    \item \textbf{Novelty} — Average of idea originality and interaction originality; \\
    \phantom{--} 10 = reveals counterintuitive second-/third-order effects 
    \item \textbf{Elaboration} — 10 = nodes have actors, dollar figures, feedback loops, depth; \\
    \phantom{--} -1 for every 3 generic nodes
    \item \textbf{Actionable} — 10 = cites sectors/tickers with entry/exit or hedge triggers; \\
    \phantom{--} 8 = clear long/short sector; $\leq 5$ = not tradable
\end{itemize}

\bigskip
\textbf{4. Fatal-flaw CAP (before scoring)}
Set every dimension to \textbf{3 or below} if \emph{any} branch contains:
\begin{itemize}
    \item Impossible outcome (e.g., Fed adopts gold standard in 2025)  
    \item Date reversal (effect precedes cause)  
    \item Verbatim repetition of >30\% of scenario paths  
\end{itemize}

\bigskip
\textbf{Example Scenario:} \\
German auto giant Volkswagen on Wednesday reported a substantial drop in first-quarter profit... Uncertainty looms around U.S. auto tariffs...

\bigskip
\textbf{Example Output:}
\begin{flushleft}
\scriptsize
\begin{verbatim}
{
  "id": "T0",
  "title": "Initial Event",
  "description": "German inflation data released at 2.2%",
  "children": [
    {
      "id": "T1A1",
      "title": "ECB maintains dovish stance",
      "date_range": "May",
      "description": "ECB maintains dovish stance, keeps rates low",
      "children": [
        {
          "id": "T1A1.1",
          "title": "German banks increase lending",
          "description": "German banks benefit from low rates, increase lending"
        },
        {
          "id": "T1A1.2",
          "title": "Euro weakens",
          "description": "Euro weakens, boosting exports"
        },
        {
          "id": "T1A1.3",
          "title": "Moderate industrial growth",
          "description": "German industry reports moderate growth"
        },
        {
          "id": "T1A1.4",
          "title": "Bundesbank caution",
          "description": "Bundesbank signals caution on inflation",
          "children": [
            {
              "id": "T2A1",
              "title": "Industry adapts to inflation",
              "date_range": "June–August",
              "description": "Companies pass on higher costs; unions request wage hikes; consumer shifts emerge",
              "children": [
                {
                  "id": "T3A1",
                  "title": "Markets stabilize",
                  "date_range": "Q3–Q4",
                  "description": "Eurozone growth outlook improves; German stocks outperform",
                  "children": [
                    {
                      "id": "T4A1",
                      "title": "Global growth stabilizes",
                      "date_range": "Q4–2025",
                      "description": "Global trade rebounds; emerging markets benefit",
                      "children": [
                        {
                          "id": "T5A1",
                          "title": "New normal",
                          "date_range": "2025+",
                          "description": "Stable growth across EU/US; ECB remains dovish"
                        }
                      ]
                    }
                  ]
                }
              ]
            }
          ]
        }
      ]
    },
    {
      "id": "T1A2",
      "title": "ECB tightens monetary policy",
      "date_range": "May",
      "description": "ECB tightens policy in response to inflation",
      "children": [
        {
          "id": "T1A2.1",
          "title": "Euro strengthens",
          "description": "Euro strengthens, hurting exports"
        },
        {
          "id": "T1A2.2",
          "title": "Industrial slowdown",
          "description": "German industry reports slower growth"
        },
        {
          "id": "T1A2.3",
          "title": "Bundesbank warns",
          "description": "Bundesbank warns of inflation risks"
        },
        {
          "id": "T1A2.4",
          "title": "Consumers cut back",
          "description": "German consumers cut back on spending",
          "children": [
            {
              "id": "T2A2",
              "title": "Industry opposes ECB policy",
              "date_range": "June–August",
              "description": "Job cuts, political backlash, delayed consumption",
              "children": [
                {
                  "id": "T3A2",
                  "title": "Markets unstable",
                  "date_range": "Q3–Q4",
                  "description": "Growth outlook declines; ECB remains hawkish",
                  "children": [
                    {
                      "id": "T4A2",
                      "title": "Global recession risk rises",
                      "date_range": "Q4–2025",
                      "description": "Trade slows; emerging markets struggle",
                      "children": [
                        {
                          "id": "T5A2",
                          "title": "Global downturn",
                          "date_range": "2025+",
                          "description": "Recession conditions persist; ECB maintains rate hikes"
                        }
                      ]
                    }
                  ]
                }
              ]
            }
          ]
        }
      ]
    }
  ]
}
\end{verbatim}
\end{flushleft}

\textbf{[Scores in JSON]} \{\texttt{"Plausibility": X, "Novelty": Y, "Elaboration": Z, "Actionable": W}\}

\textbf{[Scenario]} \{\}

\textbf{[Timeline]} \{\}

\textbf{[Scores in JSON]} \{\}
\end{tcolorbox}

\section{Appendix II - Easy and Difficult Examples for Divergent Thinking Benchmark}
\subsection*{Novelty}
\label{appendix:novelty-examples}
\begin{tcolorbox}[
  colback=gray!5!white,
  colframe=black!75!white,
  title=Easiest Scenario and Model Scores,
  fonttitle=\bfseries,
  coltitle=white,
  arc=2mm,
  boxrule=0.4pt,
  breakable
]

\textbf{\faIcon{clipboard}\quad Scenario}

The Continuous Glucose Monitoring (CGM) market is valued at 
21 billion by 2029, driven by technological advancements, greater accessibility, and the growing availability of non-invasive, over-the-counter (OTC) solutions. Key players in the market include Dexcom, Abbott, Medtronic, and Roche. The market is characterized by a surge in patent filings, with ongoing litigation trends highlighting the high stakes in this rapidly evolving market. The CGM market is expected to be driven by innovations in CGM technology, including non-invasive glucose monitoring and seamless integration with mobile applications. The emergence of OTC CGM products is broadening access to diabetes care, particularly for non-insulin-dependent patients and wellness-focused individuals. The report provides in-depth analysis of these factors and offers actionable insights for companies, investors, and healthcare leaders navigating the CGM space.

\medskip
\textbf{\faChartBar\quad Assessment Scores}
\begin{scriptsize}
\begin{tabular}{|l|c|c|c|c|c|}
\hline
\textbf{Model} & \textbf{Plausibility} & \textbf{Novelty} & \textbf{Elaboration} & \textbf{Actionable} & \textbf{Richness} \\
\hline
phi4 & 9 & 8.0 & 9 & 7 & 5.97 \\
mistral\_lg & 9 & 8.0 & 9 & 7 & 6.79 \\
llama3\_70b & 9 & 8.0 & 9 & 8 & 5.51 \\
o4\_mini & 9 & 8.0 & 10 & 9 & 7.63 \\
deepseek\_r1 & 9 & 8.5 & 9 & 9 & 4.73 \\
llama3\_8b & 8 & 6.0 & 7 & 5 & 5.75 \\
gpt4o & 9 & 8.0 & 9 & 7 & 5.57 \\
mistral\_sm & 9 & 8.0 & 9 & 8 & 5.75 \\
llama4\_scout & 8 & 7.0 & 6 & 5 & 5.43 \\
cohere\_command\_r & 8 & 7.0 & 9 & 6 & 6.16 \\
cohere\_command\_a & 9 & 8.0 & 9 & 8 & 5.72 \\
o1 & 8 & 7.0 & 9 & 8 & 5.23 \\
llama4\_maverick & 8 & 6.0 & 7 & 5 & 5.55 \\
o1\_mini & 9 & 8.0 & 10 & 8 & 7.35 \\
\hline
\end{tabular}
\end{scriptsize}

\vspace{1em}

\textbf{\faClipboard\quad Scenario}

The Industrial Protective Clothing Fabrics Market is expected to grow at a compound annual growth rate (CAGR) of 8.1\% from 2024 to 2034, reaching a valuation of US\$ 24,747.6 million by 2034. The market is driven by increased legislation related to workplace safety, industrial goods, and growing awareness related to occupational hazards. Flame-resistant and chemical-resistant fabrics are in high demand in industries such as manufacturing, construction, and oil \& gas. Key players in the industry are focusing on developing sustainable materials, such as bio-based chemicals and recycled fibers, to reduce environmental impact.

Aramid \& blends dominate the market with a 42.8\% share in the material type segment, followed by polyolefin \& blends. The oil \& gas industry holds a 29.1\% share in the end-use industry segment. The market is expected to grow rapidly in East Asia with a CAGR of 9.0\%. Key players in the market include 3M, Ansell Limited, Carrington Textiles, DuPont, Evonik Industries, and others.

The market is expected to shift towards the use of sustainable materials, with regulatory standards that manufacturers must attain. For instance, DuPont has launched Nomex Comfort with EcoForce industrial goods, a flame-resistant fabric with a bio-based chemical-repellent finish. Toray Industries, Inc. has developed the LIVMOATM 4500AS, a disposable protective garment that complies with the JIS T 8115 Type 4: Spray-tight chemical protective standard. Hazchem Safety has launched the ORKA ECO range coverall, which reduces water consumption by more than 55 liters per unit and carbon footprints by 3.7 kg CO\textsubscript{2}e per coverall.

The market is expected to grow across major regions, including North America, Latin America, Western Europe, Eastern Europe, East Asia, South Asia, and Pacific, Middle East \& Africa. The market is expected to reach a valuation of US\$ 24,747.6 million by 2034, with a CAGR of 8.1\%.

\medskip
\textbf{\faChartBar\quad Assessment Scores}
\begin{scriptsize}
\begin{tabular}{|l|c|c|c|c|c|}
\hline
\textbf{Model} & \textbf{Plausibility} & \textbf{Novelty} & \textbf{Elaboration} & \textbf{Actionable} & \textbf{Richness} \\
\hline
phi4 & 9 & 8.0 & 9 & 8 & 5.26 \\
mistral\_lg & 8 & 6.0 & 7 & 4 & 5.98 \\
llama3\_70b & 8 & 6.0 & 7 & 7 & 5.57 \\
o4\_mini & 9 & 8.0 & 9 & 9 & 7.06 \\
deepseek\_r1 & 9 & 8.5 & 9 & 9 & 5.01 \\
llama3\_8b & 7 & 6.0 & 5 & 4 & 5.94 \\
gpt4o & 9 & 8.0 & 9 & 7 & 5.48 \\
mistral\_sm & 7 & 6.0 & 8 & 7 & 6.29 \\
llama4\_scout & 9 & 8.0 & 9 & 7 & 5.81 \\
cohere\_command\_r & 7 & 6.0 & 8 & 5 & 6.24 \\
cohere\_command\_a & 9 & 9.0 & 10 & 9 & 6.09 \\
o1 & 9 & 8.0 & 9 & 8 & 4.95 \\
llama4\_maverick & 8 & 7.0 & 8 & 6 & 5.71 \\
o1\_mini & 9 & 8.0 & 10 & 7 & 7.31 \\
\hline
\end{tabular}
\end{scriptsize}

\bigskip

\textbf{\faClipboard\quad Scenario}

The 2024 Global Motor Insurance Market Report provides in-depth analysis of the global and regional motor insurance industry. Key highlights include:

\begin{itemize}
  \item Historical values for the global and regional motor insurance industry from 2020–2024 and projected figures for 2025–2029.
  \item In-depth analysis of regional markets, including rankings, premiums, and market share of top global and regional motor insurers.
  \item Comparative analysis of leading motor insurance providers, including State Farm, Progressive, PICC, Berkshire Hathaway, Ping An, and others.
  \item Key market trends, regulatory trends, and updates on the industry.
  \item Insight into key technological developments impacting the global motor insurance industry.
  \item Analysis of the competitive landscape and top 20 regional markets.
\end{itemize}

The report provides information on key performance indicators such as written premium and claims during the review period and forecast period. It also offers a detailed analysis of the regional motor insurance industry and market forecasts to 2029.

Reasons to buy the report include making strategic business decisions using in-depth historic and forecast market data, understanding key dynamics, trends, and growth opportunities in the global and regional motor insurance industry, and identifying key regulatory developments impacting market growth.

\medskip
\textbf{\faChartBar\quad Assessment Scores}
\begin{scriptsize}
\begin{tabular}{|l|c|c|c|c|c|}
\hline
\textbf{Model} & \textbf{Plausibility} & \textbf{Novelty} & \textbf{Elaboration} & \textbf{Actionable} & \textbf{Richness} \\
\hline
phi4 & 9 & 8.0 & 9 & 8 & 5.72 \\
mistral\_lg & 8 & 7.0 & 9 & 6 & 5.68 \\
llama3\_70b & 9 & 8.0 & 9 & 7 & 5.68 \\
o4\_mini & 9 & 8.0 & 9 & 8 & 7.14 \\
deepseek\_r1 & 9 & 8.5 & 9 & 9 & 5.01 \\
llama3\_8b & 6 & 5.0 & 4 & 3 & 6.09 \\
gpt4o & 9 & 8.0 & 9 & 7 & 5.83 \\
mistral\_sm & 8 & 6.0 & 7 & 7 & 5.79 \\
llama4\_scout & 8 & 6.0 & 7 & 6 & 5.57 \\
cohere\_command\_r & 9 & 7.0 & 8 & 6 & 6.10 \\
cohere\_command\_a & 9 & 8.0 & 9 & 8 & 5.28 \\
o1 & 9 & 8.5 & 9 & 8 & 6.22 \\
llama4\_maverick & 8 & 6.0 & 7 & 5 & 5.40 \\
o1\_mini & 9 & 8.0 & 9 & 7 & 7.07 \\
\hline
\end{tabular}
\end{scriptsize}
\end{tcolorbox}

\begin{tcolorbox}[
  colback=gray!5!white,
  colframe=black!75!white,
  title=Hardest Scenarios and Model Scores,
  fonttitle=\bfseries,
  coltitle=white,
  arc=2mm,
  boxrule=0.4pt,
  breakable
]

\textbf{\faClipboard\quad Scenario}

Financial services could be one of the top-performing sectors in 2025 due to the incoming administration's likely focus on deregulation. Ares Capital, a business development company, offers a forward dividend yield of 1.18\% and has delivered exceptional total returns over the long term.

President-elect Trump's pledge to increase domestic oil and gas production could benefit companies that operate pipelines, such as Enbridge, Energy Transfer, Enterprise Products Partners, and Williams Companies. Enterprise Products Partners offers the highest forward dividend yield of 6.59\% and has increased its distribution for 26 consecutive years.

Utility stocks, such as Brookfield Infrastructure, Brookfield Renewable, Dominion Energy, Evergy, and UGI, typically pay attractive dividends. Dominion Energy was one of the 2024 winners that could extend its momentum in the new year due to its presence in Northern Virginia, a global hotspot for data centers.

Big pharma stocks, such as AbbVie and Pfizer, have been longtime favorites for income investors. Pfizer's yield stands near the top of the list, but its share price has plunged more than 50\% since late 2021. The company has multiple new products on the market and a promising pipeline with 108 clinical programs.

Real estate investment trusts (REITs), such as National Storage Affiliates Trust and Realty Income, offer highly attractive dividend yields. Realty Income is one of the best REITs on the market, with close to 15,500 properties in eight countries and a 30-year history of increasing its dividend.

Retailers, such as Lowe's Companies and Target, are also Dividend Kings, with a history of increasing their dividend for at least 50 consecutive years. Target has increased its dividend for 52 consecutive years and has a strong digital channel and loyalty program.

The final four dividend stocks to buy and hold in 2025 are a mixed assortment, including Honda Motor, PepsiCo, United Parcel Service, and Verizon Communications. UPS could be poised for a strong rebound after falling almost 20\% in 2024, as the company is growing revenue and profits and focusing on higher-margin businesses.

\medskip
\textbf{\faChartBar\quad Assessment Scores}
\begin{scriptsize}
\begin{tabular}{|l|c|c|c|c|c|}
\hline
\textbf{Model} & \textbf{Plausibility} & \textbf{Novelty} & \textbf{Elaboration} & \textbf{Actionable} & \textbf{Richness} \\
\hline
phi4 & 9 & 8.0 & 9 & 8 & 5.82 \\
mistral\_lg & 6 & 4.0 & 5 & 3 & 6.35 \\
llama3\_70b & 6 & 4.0 & 5 & 5 & 5.98 \\
o4\_mini & 8 & 6.0 & 9 & 8 & 6.49 \\
deepseek\_r1 & 9 & 8.5 & 9 & 9 & 5.14 \\
llama3\_8b & 4 & 3.0 & 5 & 4 & 6.12 \\
gpt4o & 8 & 6.0 & 7 & 6 & 5.53 \\
mistral\_sm & 6 & 5.0 & 7 & 6 & 5.72 \\
llama4\_scout & 7 & 5.0 & 6 & 4 & 5.06 \\
cohere\_command\_r & 6 & 4.0 & 5 & 5 & 5.47 \\
cohere\_command\_a & 8 & 6.0 & 7 & 8 & 5.77 \\
o1 & 8 & 6.0 & 9 & 9 & 5.60 \\
llama4\_maverick & 6 & 4.0 & 5 & 3 & 5.71 \\
o1\_mini & 8 & 6.0 & 9 & 7 & 6.68 \\
\hline
\end{tabular}
\end{scriptsize}

\bigskip

\textbf{\faClipboard\quad Scenario}

The article discusses the investing origin stories of several Motley Fool analysts, including Alicia Alfiere, Bill Barker, David Meier, and Ricky Mulvey. They share their experiences and advice for getting started with investing.

Alicia Alfiere started investing in high school with the help of her father, who was initially hesitant but eventually agreed to invest in a mutual fund. She recommends reading widely and seeking out diverse perspectives to increase knowledge and understanding of investing.

Bill Barker suggests that investors learn from the greats, particularly Warren Buffett, and recommends reading his letters to shareholders and articles. He also advises starting with index funds and diversifying investments.

David Meier emphasizes the importance of patience and holding onto investments for the long term, citing his own experience of selling Meta too soon and missing out on gains. He recommends writing down the purpose of buying a stock and returning to that thesis when considering selling.

Ricky Mulvey shares his experience of investing in InvenSense, a company that was eventually overtaken by competition. He advises investors to be aware of the competitive advantage of a company and to diversify beyond what's gone up in the recent past.

The analysts also discuss the importance of community and seeking out mentors, as well as the value of journaling and understanding the emotional aspects of investing. They recommend having a holding period of 5–10 years and not being swayed by short-term market fluctuations.

The article concludes with a message of encouragement for new investors, urging them to start slow, build their community, and stay within their circle of competence. It also promotes the Motley Fool's flagship investing service, Stock Advisor.

\medskip
\textbf{\faChartBar\quad Assessment Scores}
\begin{scriptsize}
\begin{tabular}{|l|c|c|c|c|c|}
\hline
\textbf{Model} & \textbf{Plausibility} & \textbf{Novelty} & \textbf{Elaboration} & \textbf{Actionable} & \textbf{Richness} \\
\hline
phi4 & 8 & 7.0 & 6 & 5 & 4.93 \\
mistral\_lg & 6 & 5.0 & 4 & 3 & 6.14 \\
llama3\_70b & 4 & 3.0 & 5 & 2 & 5.85 \\
o4\_mini & 7 & 5.0 & 6 & 4 & 5.97 \\
deepseek\_r1 & 9 & 8.5 & 9 & 9 & 4.75 \\
llama3\_8b & 4 & 3.0 & 3 & 2 & 6.02 \\
gpt4o & 8 & 6.0 & 7 & 5 & 5.63 \\
mistral\_sm & 6 & 5.0 & 7 & 4 & 5.85 \\
llama4\_scout & 6 & 5.0 & 4 & 3 & 4.81 \\
cohere\_command\_r & 4 & 3.0 & 5 & 2 & 5.82 \\
cohere\_command\_a & 8 & 6.0 & 7 & 5 & 6.21 \\
o1 & 8 & 7.0 & 9 & 6 & 6.04 \\
llama4\_maverick & 6 & 4.0 & 5 & 3 & 4.97 \\
o1\_mini & 6 & 4.0 & 7 & 5 & 6.52 \\
\hline
\end{tabular}
\end{scriptsize}

\bigskip

\textbf{\faClipboard\quad Scenario}

SCHD shares have fallen nearly 10\% in a month. The top 10 holdings of the Schwab US Dividend Equity ETF include Pfizer, Coca-Cola, and AbbVie. Analyst expectations for these holdings are not specified in the article. The article mentions the stock prices of the mentioned companies as of Dec.\ 30, 2024, with Pfizer at 0.56, and AbbVie at \$1.30.

\medskip
\textbf{\faChartBar\quad Assessment Scores}

\begin{scriptsize}
\begin{tabular}{|l|c|c|c|c|c|}
\hline
\textbf{Model} & \textbf{Plausibility} & \textbf{Novelty} & \textbf{Elaboration} & \textbf{Actionable} & \textbf{Richness} \\
\hline
phi4 & 8 & 7 & 6 & 7 & 6.09 \\
mistral\_lg & 3 & 4 & 5 & 3 & 6.09 \\
llama3\_70b & 3 & 2 & 4 & 3 & 5.93 \\
o4\_mini & 3 & 4 & 6 & 3 & 6.27 \\
deepseek\_r1 & 8 & 9 & 9 & 10 & 4.78 \\
llama3\_8b & 3 & 2 & 4 & 3 & 6.09 \\
gpt4o & 4 & 5 & 6 & 3 & 5.51 \\
mistral\_sm & 3 & 4 & 5 & 3 & 6.97 \\
llama4\_scout & 3 & 4 & 5 & 3 & 5.63 \\
cohere\_command\_r & 3 & 4 & 5 & 3 & 5.17 \\
cohere\_command\_a & 8 & 9 & 9 & 10 & 5.52 \\
o1 & 3 & 4 & 5 & 2 & 5.56 \\
llama4\_maverick & 3 & 4 & 5 & 3 & 5.29 \\
o1\_mini & 3 & 4 & 6 & 2 & 6.60 \\
\hline
\end{tabular}
\end{scriptsize}
\end{tcolorbox}

\subsection*{Elaboration}
\label{appendix:elaboration-examples}
\begin{tcolorbox}[
  colback=gray!5!white,
  colframe=black!75!white,
  title=Easiest Scenarios and Model Scores,
  fonttitle=\bfseries,
  coltitle=white,
  arc=2mm,
  boxrule=0.4pt,
  breakable
]

\textbf{\faClipboard\quad Scenario}

The Continuous Glucose Monitoring (CGM) market is valued at \$21 billion by 2029, driven by technological advancements, greater accessibility, and the growing availability of non-invasive, over-the-counter (OTC) solutions. Key players in the market include Dexcom, Abbott, Medtronic, and Roche. The market is characterized by a surge in patent filings, with ongoing litigation trends highlighting the high stakes in this rapidly evolving market. The CGM market is expected to be driven by innovations in CGM technology, including non-invasive glucose monitoring and seamless integration with mobile applications. The emergence of OTC CGM products is broadening access to diabetes care, particularly for non-insulin-dependent patients and wellness-focused individuals. The report provides in-depth analysis of these factors and offers actionable insights for companies, investors, and healthcare leaders navigating the CGM space.

\medskip
\textbf{\faChartBar\quad Assessment Scores}

\begin{scriptsize}
\begin{tabular}{|l|c|c|c|c|c|}
\hline
\textbf{Model} & \textbf{Plausibility} & \textbf{Novelty} & \textbf{Elaboration} & \textbf{Actionable} & \textbf{Richness} \\
\hline
phi4 & 9 & 8.0 & 9 & 7 & 5.97 \\
mistral\_lg & 9 & 8.0 & 9 & 7 & 6.79 \\
llama3\_70b & 9 & 8.0 & 9 & 8 & 5.51 \\
o4\_mini & 9 & 8.0 & 10 & 9 & 7.63 \\
deepseek\_r1 & 9 & 8.5 & 9 & 9 & 4.73 \\
llama3\_8b & 8 & 6.0 & 7 & 5 & 5.75 \\
gpt4o & 9 & 8.0 & 9 & 7 & 5.57 \\
mistral\_sm & 9 & 8.0 & 9 & 8 & 5.75 \\
llama4\_scout & 8 & 7.0 & 6 & 5 & 5.43 \\
cohere\_command\_r & 8 & 7.0 & 9 & 6 & 6.16 \\
cohere\_command\_a & 9 & 8.0 & 9 & 8 & 5.72 \\
o1 & 8 & 7.0 & 9 & 8 & 5.23 \\
llama4\_maverick & 8 & 6.0 & 7 & 5 & 5.55 \\
o1\_mini & 9 & 8.0 & 10 & 8 & 7.35 \\
\hline
\end{tabular}
\end{scriptsize}

\bigskip

\textbf{\faClipboard\quad Scenario}

The global hydrogen market was valued at US\$1,657.24 billion by 2050, growing at a CAGR of 7.88\% during the forecast period 2025–2050. The market is driven by the urgency to decarbonize and the emergence of powerful policy incentives. Over 1,400 new hydrogen production projects were announced worldwide from 2020 to 2023, with Europe accounting for over half of the world's announced renewable hydrogen projects.

\textbf{Key Market Findings:}
\begin{itemize}
  \item Forecast (2050): US\$1,657.24 billion
  \item CAGR: 7.88\%
  \item Largest Region (2024): Asia Pacific (45.83\%)
  \item Type: Grey Hydrogen (82.13\%), Technology: Thermal Process (86.79\%)
  \item Delivery Mode: Merchant (55.43\%), Application: Ammonia Production (36.04\%)
\end{itemize}

\textbf{Top Drivers:}
\begin{itemize}
  \item Accelerating policy support
  \item Falling renewables reducing green hydrogen costs
  \item Rising demand for decarbonized feedstock
\end{itemize}

\textbf{Top Trends:}
\begin{itemize}
  \item Growing cross-border hydrogen trade
  \item Electrolyzer innovations reducing capital costs
  \item Integration of blended hydrogen systems across sectors
\end{itemize}

\textbf{Top Challenges:}
\begin{itemize}
  \item High production costs
  \item Infrastructure constraints
  \item Inconsistent policy frameworks
\end{itemize}

Global hydrogen investments are estimated at US\$9 trillion by 2050 to meet net-zero goals. Major companies include Air Liquide, Plug Power, Siemens AG, Cummins, Linde, Enapter AG, and others.

\medskip
\textbf{\faChartBar\quad Assessment Scores}

\begin{scriptsize}
\begin{tabular}{|l|c|c|c|c|c|}
\hline
\textbf{Model} & \textbf{Plausibility} & \textbf{Novelty} & \textbf{Elaboration} & \textbf{Actionable} & \textbf{Richness} \\
\hline
phi4 & 9 & 8 & 8 & 7 & 5.48 \\
mistral\_lg & 8 & 6 & 7 & 5 & 6.33 \\
llama3\_70b & 8 & 6 & 7 & 5 & 5.83 \\
o4\_mini & 8 & 7 & 9 & 6 & 7.45 \\
deepseek\_r1 & 9 & 9 & 10 & 9 & 5.02 \\
llama3\_8b & 6 & 5 & 7 & 4 & 6.67 \\
gpt4o & 9 & 8 & 9 & 7 & 5.87 \\
mistral\_sm & 8 & 6 & 7 & 5 & 6.82 \\
llama4\_scout & 8 & 7 & 9 & 6 & 5.94 \\
cohere\_command\_r & 8 & 7 & 9 & 6 & 6.56 \\
cohere\_command\_a & 9 & 8 & 9 & 9 & 6.02 \\
o1 & 8 & 7 & 9 & 7 & 5.77 \\
llama4\_maverick & 9 & 8 & 9 & 8 & 5.91 \\
o1\_mini & 8 & 7 & 9 & 6 & 7.64 \\
\hline
\end{tabular}
\end{scriptsize}

\bigskip

\textbf{\faClipboard\quad Scenario}

The Industrial Protective Clothing Fabrics Market is expected to grow at a CAGR of 8.1\% from 2024 to 2034, reaching a valuation of US\$24,747.6 million. Growth is driven by workplace safety regulations and increased awareness of occupational hazards. High demand persists in manufacturing, construction, and oil \& gas sectors for flame- and chemical-resistant fabrics.

\textbf{Key Highlights:}
\begin{itemize}
  \item Aramid \& blends dominate material segment (42.8\%)
  \item Oil \& gas leads end-use sector (29.1\%)
  \item East Asia projected highest CAGR (9.0\%)
\end{itemize}

\textbf{Sustainability Trends:}
\begin{itemize}
  \item DuPont’s Nomex Comfort with EcoForce
  \item Toray’s LIVMOA 4500AS, a disposable garment
  \item ORKA ECO by Hazchem, reducing water and carbon
\end{itemize}

\textbf{Major Regions:} North America, Latin America, Europe, Asia Pacific, Middle East \& Africa.

\medskip
\textbf{\faChartBar\quad Assessment Scores}

\begin{scriptsize}
\begin{tabular}{|l|c|c|c|c|c|}
\hline
\textbf{Model} & \textbf{Plausibility} & \textbf{Novelty} & \textbf{Elaboration} & \textbf{Actionable} & \textbf{Richness} \\
\hline
phi4 & 9 & 8.0 & 9 & 8 & 5.26 \\
mistral\_lg & 8 & 6.0 & 7 & 4 & 5.98 \\
llama3\_70b & 8 & 6.0 & 7 & 7 & 5.57 \\
o4\_mini & 9 & 8.0 & 9 & 9 & 7.06 \\
deepseek\_r1 & 9 & 8.5 & 9 & 9 & 5.01 \\
llama3\_8b & 7 & 6.0 & 5 & 4 & 5.94 \\
gpt4o & 9 & 8.0 & 9 & 7 & 5.48 \\
mistral\_sm & 7 & 6.0 & 8 & 7 & 6.29 \\
llama4\_scout & 9 & 8.0 & 9 & 7 & 5.81 \\
cohere\_command\_r & 7 & 6.0 & 8 & 5 & 6.24 \\
cohere\_command\_a & 9 & 9.0 & 10 & 9 & 6.09 \\
o1 & 9 & 8.0 & 9 & 8 & 4.95 \\
llama4\_maverick & 8 & 7.0 & 8 & 6 & 5.71 \\
o1\_mini & 9 & 8.0 & 10 & 7 & 7.31 \\
\hline
\end{tabular}
\end{scriptsize}

\end{tcolorbox}

\begin{tcolorbox}[
  colback=gray!5!white,
  colframe=black!75!white,
  title=Hardest Scenarios and Model Scores,
  fonttitle=\bfseries,
  coltitle=white,
  arc=2mm,
  boxrule=0.4pt,
  breakable
]

\textbf{\faClipboard\quad Scenario}

SCHD shares have fallen nearly 10\% in a month. The ETF's top 10 holdings include Pfizer, Coca-Cola, and AbbVie. Analyst expectations for these holdings are not specified. The article mentions a video discussing the pros and cons of SCHD and analyst expectations for its top holdings, but does not provide any information from the video.

Coca-Cola's stock price is 0.56\% of the ETF's total, and its current stock price is not specified.

\medskip
\textbf{\faChartBar\quad Assessment Scores}

\begin{scriptsize}
\begin{tabular}{|l|c|c|c|c|c|}
\hline
\textbf{Model} & \textbf{Plausibility} & \textbf{Novelty} & \textbf{Elaboration} & \textbf{Actionable} & \textbf{Richness} \\
\hline
phi4 & 8 & 7 & 6 & 5 & 6.29 \\
mistral\_lg & 6 & 4 & 5 & 3 & 5.79 \\
llama3\_70b & 6 & 5 & 4 & 3 & 6.09 \\
o4\_mini & 7 & 6 & 8 & 5 & 6.77 \\
llama3\_8b & 4 & 3 & 5 & 3 & 5.94 \\
gpt4o & 7 & 6 & 6 & 5 & 5.51 \\
mistral\_sm & 8 & 7 & 6 & 5 & 6.29 \\
llama4\_scout & 6 & 5 & 4 & 3 & 5.28 \\
cohere\_command\_r & 4 & 3 & 5 & 3 & 5.83 \\
cohere\_command\_a & 9 & 8 & 9 & 9 & 5.70 \\
o1 & 8 & 7 & 9 & 6 & 5.93 \\
llama4\_maverick & 7 & 5 & 6 & 4 & 5.63 \\
o1\_mini & 8 & 6 & 7 & 6 & 6.85 \\
\hline
\end{tabular}
\end{scriptsize}

\textbf{\faClipboard\quad Scenario}

The article discusses the investing origin stories of several Motley Fool analysts, including Alicia Alfiere, Bill Barker, David Meier, and Ricky Mulvey. They share their experiences and advice for getting started with investing.

Alicia Alfiere started investing in high school with the help of her father, who was initially hesitant but eventually agreed to invest in a mutual fund. She recommends reading widely and seeking out diverse perspectives to increase knowledge and understanding of investing.

Bill Barker suggests that investors learn from the greats, particularly Warren Buffett, and recommends reading his letters to shareholders and articles. He also advises starting with index funds and diversifying investments.

David Meier emphasizes the importance of patience and holding onto investments for the long term, citing his own experience of selling Meta too soon and missing out on gains. He recommends writing down the purpose of buying a stock and returning to that thesis when considering selling.

Ricky Mulvey shares his experience of investing in InvenSense, a company that was eventually overtaken by competition. He advises investors to be aware of the competitive advantage of a company and to diversify beyond what's gone up in the recent past.

The analysts also discuss the importance of community and seeking out mentors, as well as the value of journaling and understanding the emotional aspects of investing. They recommend having a holding period of 5--10 years and not being swayed by short-term market fluctuations.

\medskip
\textbf{\faChartBar\quad Assessment Scores}

\begin{scriptsize}
\begin{tabular}{|l|c|c|c|c|c|}
\hline
\textbf{Model} & \textbf{Plausibility} & \textbf{Novelty} & \textbf{Elaboration} & \textbf{Actionable} & \textbf{Richness} \\
\hline
phi4 & 8 & 7.0 & 6 & 5 & 4.93 \\
mistral\_lg & 6 & 5.0 & 4 & 3 & 6.14 \\
llama3\_70b & 4 & 3.0 & 5 & 2 & 5.85 \\
o4\_mini & 7 & 5.0 & 6 & 4 & 5.97 \\
deepseek\_r1 & 9 & 8.5 & 9 & 9 & 4.75 \\
llama3\_8b & 4 & 3.0 & 3 & 2 & 6.02 \\
gpt4o & 8 & 6.0 & 7 & 5 & 5.63 \\
mistral\_sm & 6 & 5.0 & 7 & 4 & 5.85 \\
llama4\_scout & 6 & 5.0 & 4 & 3 & 4.81 \\
cohere\_command\_r & 4 & 3.0 & 5 & 2 & 5.82 \\
cohere\_command\_a & 8 & 6.0 & 7 & 5 & 6.21 \\
o1 & 8 & 7.0 & 9 & 6 & 6.04 \\
llama4\_maverick & 6 & 4.0 & 5 & 3 & 4.97 \\
o1\_mini & 6 & 4.0 & 7 & 5 & 6.52 \\
\hline
\end{tabular}
\end{scriptsize}

\textbf{\faClipboard\quad Scenario}

SCHD shares have fallen nearly 10\% in a month. The top 10 holdings of the Schwab US Dividend Equity ETF include Pfizer, Coca-Cola, and AbbVie. Analyst expectations for these holdings are not specified in the article. The article mentions the stock prices of the mentioned companies as of Dec.\ 30, 2024, with Pfizer at 0.56, and AbbVie at \$1.30.

\medskip
\textbf{\faChartBar\quad Assessment Scores}

\begin{scriptsize}
\begin{tabular}{|l|c|c|c|c|c|}
\hline
\textbf{Model} & \textbf{Plausibility} & \textbf{Novelty} & \textbf{Elaboration} & \textbf{Actionable} & \textbf{Richness} \\
\hline
phi4 & 8 & 7 & 6 & 7 & 6.09 \\
mistral\_lg & 3 & 4 & 5 & 3 & 6.09 \\
llama3\_70b & 3 & 2 & 4 & 3 & 5.93 \\
o4\_mini & 3 & 4 & 6 & 3 & 6.27 \\
deepseek\_r1 & 8 & 9 & 9 & 10 & 4.78 \\
llama3\_8b & 3 & 2 & 4 & 3 & 6.09 \\
gpt4o & 4 & 5 & 6 & 3 & 5.51 \\
mistral\_sm & 3 & 4 & 5 & 3 & 6.97 \\
llama4\_scout & 3 & 4 & 5 & 3 & 5.63 \\
cohere\_command\_r & 3 & 4 & 5 & 3 & 5.17 \\
cohere\_command\_a & 8 & 9 & 9 & 10 & 5.52 \\
o1 & 3 & 4 & 5 & 2 & 5.56 \\
llama4\_maverick & 3 & 4 & 5 & 3 & 5.29 \\
o1\_mini & 3 & 4 & 6 & 2 & 6.60 \\
\hline
\end{tabular}
\end{scriptsize}

\end{tcolorbox}

\subsection*{Actionable}
\label{appendix:actionable-examples}

\begin{tcolorbox}[
  colback=gray!5!white,
  colframe=black!75!white,
  title=Easiest Scenarios and Model Scores,
  fonttitle=\bfseries,
  coltitle=white,
  arc=2mm,
  boxrule=0.4pt,
  breakable
]

\textbf{\faClipboard\quad Scenario}

JPMorgan analysts project a -0.3\% U.S. GDP contraction in 2025, with a 60\% probability of a U.S. recession, up from 40\% last month due to recent tariff hikes. Despite this, utility companies are seen as relatively well-positioned due to their regulated return profile and cost-of-service structure, which insulates them from direct tariff-related impacts.

Potential rate cuts may further boost income appeal for utilities. However, prolonged tariff uncertainty could harm sentiment and the broader economy, creating affordability challenges that may impact utilities' capex plans and returns, especially in regions with tight energy supply.

JPMorgan analyst Jeremy Tonet downgraded Dominion Energy, Inc.\ (D) from Neutral to Underweight and cut the price forecast from 52, citing increased risks for the company's Coastal Virginia Offshore Wind project due to potential tariff impacts and still-to-be-finalized PJM network upgrades.

On the other hand, Tonet raised the rating for WEC Energy Group, Inc.\ (WEC) from Underweight to Neutral and the price forecast from 108, citing the company's lack of rate case exposure and favorable regulatory environment in Wisconsin.

Tonet also upgraded Southern Company (SO) from Underweight to Neutral, citing the company's strong regulatory environment, solid balance sheet, and economic resilience in its service area. However, ongoing regulatory uncertainty limits immediate upside.

\medskip
\textbf{\faChartBar\quad Assessment Scores}

\begin{scriptsize}
\begin{tabular}{|l|c|c|c|c|c|}
\hline
\textbf{Model} & \textbf{Plausibility} & \textbf{Novelty} & \textbf{Elaboration} & \textbf{Actionable} & \textbf{Richness} \\
\hline
phi4 & 9 & 8 & 9 & 8 & 5.51 \\
mistral\_lg & 8 & 6 & 7 & 8 & 5.53 \\
llama3\_70b & 9 & 8 & 9 & 9 & 5.51 \\
o4\_mini & 8 & 7 & 9 & 8 & 6.39 \\
llama3\_8b & 8 & 6 & 7 & 8 & 5.94 \\
gpt4o & 8 & 7 & 9 & 8 & 5.82 \\
mistral\_sm & 8 & 6 & 7 & 8 & 6.13 \\
llama4\_scout & 8 & 7 & 9 & 8 & 5.46 \\
cohere\_command\_r & 7 & 6 & 8 & 7 & 5.82 \\
cohere\_command\_a & 9 & 8 & 9 & 9 & 5.83 \\
o1 & 8 & 7 & 9 & 8 & 5.87 \\
llama4\_maverick & 8 & 6 & 7 & 7 & 6.09 \\
o1\_mini & 9 & 8 & 9 & 9 & 6.96 \\
\hline
\end{tabular}
\end{scriptsize}

\textbf{\faClipboard\quad Scenario}

Tariffs are expected to have a limited impact on inflation, contrary to popular belief. A recent study by the Centre for Economic Policy Research found that tariffs do not boost inflation as they are a short-term headwind on growth. The Financial Times also reached a similar conclusion, stating that tariffs compress company margins but are typically absorbed by firms and their shareholders.

As a result, the bond market is not pricing in high inflation expectations, with the 10-year Treasury yield remaining below 5\%. This presents an opportunity for investors to buy bonds and bond proxies, such as the Cohen \& Steers Infrastructure Fund (UTF), which offers a 7.6\% yield. UTF provides instant exposure to 256 strong utility names and holds 16\% in corporate bonds, making it a smart ``rinse-and-repeat'' trade on the 10-year Treasury yield.

The 10-year Treasury yield is expected to fall further on a slowing economy, making it a good time to buy UTF. The fund trades near par and is a good opportunity to buy when it's near or below its net asset value (NAV).

\medskip
\textbf{\faChartBar\quad Assessment Scores}

\begin{scriptsize}
\begin{tabular}{|l|c|c|c|c|c|}
\hline
\textbf{Model} & \textbf{Plausibility} & \textbf{Novelty} & \textbf{Elaboration} & \textbf{Actionable} & \textbf{Richness} \\
\hline
phi4 & 9 & 8 & 9 & 9 & 6.43 \\
mistral\_lg & 7 & 6 & 8 & 7 & 5.88 \\
llama3\_70b & 6 & 5 & 5 & 7 & 5.68 \\
o4\_mini & 8 & 6 & 9 & 7 & 6.59 \\
deepseek\_r1 & 9 & 8 & 9 & 10 & 4.37 \\
llama3\_8b & 6 & 4 & 5 & 7 & 6.01 \\
gpt4o & 8 & 6 & 7 & 8 & 5.51 \\
mistral\_sm & 6 & 5 & 4 & 6 & 6.29 \\
llama4\_scout & 7 & 6 & 5 & 6 & 5.15 \\
cohere\_command\_r & 6 & 5 & 4 & 6 & 6.67 \\
cohere\_command\_a & 9 & 8 & 9 & 9 & 6.43 \\
o1 & 8 & 7 & 9 & 8 & 5.57 \\
llama4\_maverick & 7 & 6 & 5 & 7 & 5.94 \\
o1\_mini & 8 & 6 & 7 & 8 & 6.42 \\
\hline
\end{tabular}
\end{scriptsize}

\textbf{\faClipboard\quad Scenario}

Crude oil prices remained relatively stable in 2024, with Brent oil slipping 3\% and WTI ending the year at around the \$70s. As a result, oil stocks will need other catalysts to drive share prices.

ConocoPhillips (COP) and Chevron (CVX) are two oil stocks with notable catalysts. ConocoPhillips acquired Marathon Oil in a \$30 per barrel deal. The company expects to capture over \$1 billion in cost and capital synergies within the first 12 months, boosting its free cash flow. ConocoPhillips plans to return a meaningful percentage of its growing cash flow to shareholders through dividend growth and share repurchases.

Chevron is working on closing a \$60 billion all-stock deal to buy Hess, which would significantly upgrade and diversify its portfolio. The deal would enhance and extend Chevron's production and free cash flow growth outlook into the 2030s, helping it more than double its free cash flow by 2027. If Chevron wins the arbitration case, it will be able to close the acquisition, providing a big boost in 2025 and beyond. Even if it loses, Chevron expects to grow its free cash flow by more than 10\% annually through 2027, fueled by high-return capital investments into its existing resource portfolio.

\medskip
\textbf{\faChartBar\quad Assessment Scores}

\begin{scriptsize}
\begin{tabular}{|l|c|c|c|c|c|}
\hline
\textbf{Model} & \textbf{Plausibility} & \textbf{Novelty} & \textbf{Elaboration} & \textbf{Actionable} & \textbf{Richness} \\
\hline
phi4 & 9 & 8 & 9 & 9 & 5.39 \\
mistral\_lg & 8 & 6 & 7 & 6 & 6.09 \\
llama3\_70b & 8 & 6 & 7 & 8 & 5.18 \\
o4\_mini & 8 & 6 & 7 & 8 & 5.71 \\
deepseek\_r1 & 9 & 8 & 9 & 9 & 4.85 \\
llama3\_8b & 8 & 6 & 7 & 6 & 6.20 \\
gpt4o & 8 & 6 & 7 & 6 & 5.22 \\
mistral\_sm & 8 & 6 & 7 & 6 & 6.24 \\
llama4\_scout & 9 & 8 & 9 & 8 & 5.39 \\
cohere\_command\_r & 8 & 6 & 7 & 6 & 5.51 \\
cohere\_command\_a & 8 & 6 & 8 & 9 & 5.69 \\
o1 & 9 & 8 & 9 & 8 & 5.56 \\
llama4\_maverick & 8 & 6 & 7 & 7 & 5.53 \\
o1\_mini & 9 & 7 & 8 & 8 & 6.49 \\
\hline
\end{tabular}
\end{scriptsize}

\end{tcolorbox}

\begin{tcolorbox}[
  colback=gray!5!white,
  colframe=black!75!white,
  title=Hardest Scenarios and Model Scores,
  fonttitle=\bfseries,
  coltitle=white,
  arc=2mm,
  boxrule=0.4pt,
  breakable
]

\textbf{\faClipboard\quad Scenario}

The report highlights qualitative insights into investor behavior and best practices based on the personal experiences of investment analysts. Key themes emphasize the long-term benefits of early exposure to capital markets, the importance of foundational knowledge through literature from established investors such as Warren Buffett, and strategic portfolio diversification using index funds. Analysts stress the significance of patience in investment decisions, supported by case examples underscoring the risks of premature divestment. The report also underscores cognitive and emotional factors in investment performance, advocating for structured investment theses, mentor networks, and journaling to reinforce discipline. A consensus view recommends a long-term holding horizon of 5–10 years to withstand market volatility. The findings collectively support a behavioral and strategic framework for retail investors aiming to improve decision-making and resilience in dynamic market environments.

\medskip
\textbf{\faChartBar\quad Assessment Scores}

\begin{scriptsize}
\begin{tabular}{|l|c|c|c|c|c|}
\hline
\textbf{Model} & \textbf{Plausibility} & \textbf{Novelty} & \textbf{Elaboration} & \textbf{Actionable} & \textbf{Richness} \\
\hline
phi4 & 8 & 7.0 & 6 & 5 & 4.93 \\
mistral\_lg & 6 & 5.0 & 4 & 3 & 6.14 \\
llama3\_70b & 4 & 3.0 & 5 & 2 & 5.85 \\
o4\_mini & 7 & 5.0 & 6 & 4 & 5.97 \\
deepseek\_r1 & 9 & 8.5 & 9 & 9 & 4.75 \\
llama3\_8b & 4 & 3.0 & 3 & 2 & 6.02 \\
gpt4o & 8 & 6.0 & 7 & 5 & 5.63 \\
mistral\_sm & 6 & 5.0 & 7 & 4 & 5.85 \\
llama4\_scout & 6 & 5.0 & 4 & 3 & 4.81 \\
cohere\_command\_r & 4 & 3.0 & 5 & 2 & 5.82 \\
cohere\_command\_a & 8 & 6.0 & 7 & 5 & 6.21 \\
o1 & 8 & 7.0 & 9 & 6 & 6.04 \\
llama4\_maverick & 6 & 4.0 & 5 & 3 & 4.97 \\
o1\_mini & 6 & 4.0 & 7 & 5 & 6.52 \\
\hline
\end{tabular}
\end{scriptsize}

\textbf{\faClipboard\quad Scenario}

SCHD shares have fallen nearly 10\% in a month. The top 10 holdings of the Schwab US Dividend Equity ETF include Pfizer, Coca-Cola, and AbbVie. Analyst expectations for these holdings are not specified in the article. The article mentions the stock prices of the mentioned companies as of Dec. 30, 2024, with Pfizer at 0.56, and AbbVie at \$1.30.

\medskip
\textbf{\faChartBar\quad Assessment Scores}

\begin{scriptsize}
\begin{tabular}{|l|c|c|c|c|c|}
\hline
\textbf{Model} & \textbf{Plausibility} & \textbf{Novelty} & \textbf{Elaboration} & \textbf{Actionable} & \textbf{Richness} \\
\hline
phi4 & 8 & 7 & 6 & 7 & 6.09 \\
mistral\_lg & 3 & 4 & 5 & 3 & 6.09 \\
llama3\_70b & 3 & 2 & 4 & 3 & 5.93 \\
o4\_mini & 3 & 4 & 6 & 3 & 6.27 \\
deepseek\_r1 & 8 & 9 & 9 & 10 & 4.78 \\
llama3\_8b & 3 & 2 & 4 & 3 & 6.09 \\
gpt4o & 4 & 5 & 6 & 3 & 5.51 \\
mistral\_sm & 3 & 4 & 5 & 3 & 6.97 \\
llama4\_scout & 3 & 4 & 5 & 3 & 5.63 \\
cohere\_command\_r & 3 & 4 & 5 & 3 & 5.17 \\
cohere\_command\_a & 8 & 9 & 9 & 10 & 5.52 \\
o1 & 3 & 4 & 5 & 2 & 5.56 \\
llama4\_maverick & 3 & 4 & 5 & 3 & 5.29 \\
o1\_mini & 3 & 4 & 6 & 2 & 6.60 \\
\hline
\end{tabular}
\end{scriptsize}

\textbf{\faClipboard\quad Scenario}

Shiba Inu (SHIB) cryptocurrency, a meme token with no real utility, has a current price of 0.000086 in 2021. Despite its volatility, SHIB more than doubled in 2024 as investors speculated on another historic rally. The cryptocurrency is riding tailwinds including risk-on sentiment, falling interest rates, and a pro-crypto U.S. government.

The U.S. economy remains strong, and the Federal Reserve has started cutting interest rates, making growth assets like stocks and cryptocurrencies more enticing to investors. The pro-crypto administration of former President Donald Trump has also created a positive environment for speculative assets like SHIB.

However, SHIB faces fundamental challenges, including a supply problem, with 589.5 trillion tokens in circulation. To reach a price of \$1, nearly all of the current supply would need to be eliminated, which is theoretically possible but highly unlikely. Burning tokens would only reduce supply, not create value, and would take 20,460 years at the current pace.

The market capitalization of SHIB is currently \$1 would make it 159 times more valuable than Apple, the world's largest company. The Shiba Inu community is trying to solve the supply problem by burning tokens, but this would not create any value and would not change the net financial position of investors.

\medskip
\textbf{\faChartBar\quad Assessment Scores}

\begin{scriptsize}
\begin{tabular}{|l|c|c|c|c|c|}
\hline
\textbf{Model} & \textbf{Plausibility} & \textbf{Novelty} & \textbf{Elaboration} & \textbf{Actionable} & \textbf{Richness} \\
\hline
phi4 & 8 & 7 & 6 & 5 & 5.28 \\
mistral\_lg & 4 & 6 & 5 & 3 & 5.79 \\
llama3\_70b & 4 & 5 & 6 & 3 & 5.47 \\
o4\_mini & 4 & 5 & 6 & 3 & 6.60 \\
deepseek\_r1 & 8 & 9 & 9 & 8 & 4.96 \\
llama3\_8b & 4 & 3 & 5 & 2 & 5.62 \\
gpt4o & 4 & 6 & 5 & 3 & 6.03 \\
mistral\_sm & 8 & 7 & 6 & 5 & 6.09 \\
llama4\_scout & 4 & 5 & 6 & 3 & 5.68 \\
cohere\_command\_r & 4 & 6 & 7 & 3 & 7.27 \\
cohere\_command\_a & 8 & 7 & 9 & 6 & 6.67 \\
o1 & 4 & 6 & 7 & 5 & 5.47 \\
llama4\_maverick & 4 & 5 & 6 & 3 & 5.35 \\
o1\_mini & 4 & 6 & 7 & 5 & 7.02 \\
\hline
\end{tabular}
\end{scriptsize}

\end{tcolorbox}

\section{Appendix III - Convergent Thinking System and User Prompt}
\onecolumn
\subsection*{Convergent Thinking Question Generation Prompt}
\label{appendix:convergent-qn-generation-prompt-template}
\begin{tcolorbox}[
  colback=gray!5!white,
  colframe=black!75!white,
  title=Convergent Thinking Prompt: Timeline Generation,
  fonttitle=\bfseries,
  coltitle=white,
  arc=2mm,
  boxrule=0.4pt,
  breakable,
  sharp corners,
  enhanced jigsaw,
  width=\textwidth,
  before upper={\setlength{\parskip}{4pt}},
]

You are a \textbf{convergent-thinking MCQ generator} for economics/finance/politics.  
Use \textbf{all six} of these adversarial pipelines to produce \textbf{one} JSON question with \{num\_options\} timelines (1 correct, \{num\_options\}–1 hard distractors):

\begin{itemize}    
    \item \textbf{Historical $\beta$-Swap}: Swap one latent driver in a real macro event.  
    \item \textbf{Numeric Trip-Wire}: Embed a mini calc (e.g. margin math, FX pass-through).  
    \item \textbf{Policy Game}: Use a toy policy-model state → next-state logic to spawn one correct seq, one that violates a key equation.  
    \item \textbf{Cross-Section Confuser}: Include two firms with opposite exposures; only one timeline matches Firm A.  
    \item \textbf{Reg-Legal Trap}: Build a legal/reg-process chronology; distractors break one procedural step.  
    \item \textbf{Adversarial Self-Play}: Generate many distractors, judge them against the correct path, keep only those within 0.05 plausibility of the correct one.
\end{itemize}

\textbf{Instructions}:  
\begin{itemize}
    \item Read the article below \{article\}.  
    \item Sample 5-7 factors.  
    \item Invent $main\_event$ + 3-5-sentence $event\_description$ with conflicting signals.  
    \item Generate exactly $num\_options$ timelines of 5-7 steps each.  
    \item \textbf{Correct} timeline must satisfy all convergent criteria (factor alignment, temporal coherence, logical entailment).  
    \item \textbf{Distractors}: each violates exactly one convergent rule via \textbf{one} of the six pipelines above.  
    \item Annotate each timeline with "is\_correct": true/false and "reason" naming the violated rule or why it's correct.
\end{itemize}

Generate a question set based on the following article:

\#\#\# Article Start \#\#\#

\{article\}

\#\#\# Article End \#\#\#

\medskip
Output in JSON.

e.g for Company: Nvidia, with the following selected factors: Protectionism, Trade War, Economic Slowdown, Currency Devaluation, Geopolitical Rivalry

\medskip
\begin{verbatim}
{
  "company": "Nvidia",
  "main_event": "U.S Tariffs on Everything",
  "event_description": "It is January 2025. Donald Trump is re-elected and proposes a sweeping 20% tariff on 
  all foreign imports, including semiconductors, steel, and consumer electronics. While 
  the proposal sparks immediate volatility, markets initially react positively to strong Q4 earnings across tech. 
  Nvidia, which relies heavily on Taiwan and South Korea for advanced packaging and fabrication, 
  also benefits from surging AI demand. However, its executives express concern about long-term supply chain 
  exposure and cost pass-through limitations. Industry lobbying intensifies as U.S. policymakers signal possible 
  carve-outs for strategic industries. The timeline and final structure of the tariff implementation remain 
  unclear.",
  "factors": [
    "Protectionism",
    "Trade War",
    "Economic Slowdown",
    "Currency Devaluation",
    "Geopolitical Rivalry"
  ],
  "timelines": [
    {
      "path": [
        "Tariff proposal announced",
        "Semiconductor sector volatility spikes",
        "Nvidia stock initially rises on strong Q4 earnings",
        "Analysts raise concerns over supply chain exposure",
        "Lobbying efforts begin for semiconductor carve-outs",
        "Investor uncertainty grows as implementation timeline remains unclear"
      ],
      "is_correct": true,
      "reason": "Aligns with market optimism and medium-term uncertainty; preserves logical and temporal 
      coherence."
    },
    {
      "path": [
        "Tariff proposal announced",
        "Nvidia pivots to U.S. domestic chip production",
        "Stock surges 20% on investor optimism",
        "China announces retaliatory export bans",
        "Nvidia expands market share in Asia"
      ],
      "is_correct": false,
      "reason": "Violates temporal coherence—domestic production shifts take quarters; Asian market wouldn't 
      expand post-ban."
    },
    {
      "path": [
        "Tariff proposal announced",
        "AI sector demand collapses",
        "Nvidia stock plunges 30%",
        "Nvidia announces layoffs",
        "Company exits consumer GPU business"
      ],
      "is_correct": false,
      "reason": "Contradicts stated strong AI demand and includes unjustified drastic corporate moves."
    },
    {
      "path": [
        "Tariff proposal announced",
        "Foreign suppliers raise prices",
        "Nvidia absorbs increased costs",
        "Analysts upgrade Nvidia on margin resilience",
        "Stock rises steadily through Q1"
      ],
      "is_correct": false,
      "reason": "Violates factor alignment—margin pressure and geopolitical uncertainty are ignored."
    }
  ]
}
\end{verbatim}
\end{tcolorbox}

\subsection*{Convergent Thinking Question Refinement Prompt}
\label{appendix:convergent-refinement-prompt-template}
\begin{tcolorbox}[
  colback=gray!5!white,
  colframe=black!75!white,
  title=Convergent Thinking Prompt: Refine Existing Question,
  fonttitle=\bfseries,
  coltitle=white,
  arc=2mm,
  boxrule=0.4pt,
  breakable
]

You are a \textbf{reviewer for a convergent thinking assessment} designed to evaluate analytical reasoning in economic and geopolitical domains.  
Each question presents a macroeconomic or policy scenario, followed by multiple timeline options.  
Your role is to validate whether the test item meets the difficulty standard for trained analysts by checking if you can easily identify the correct option using the convergent reasoning criteria.
\bigskip
\textbf{Convergent Criteria:}
\begin{enumerate}
  \item Events must exhibit strong \textbf{entailment} from one to the next.
  \item Events must align with the \textbf{key factors} stated in the scenario.
  \item Timelines must exhibit \textbf{temporal coherence}, with no implausible jumps or accelerations.
\end{enumerate}
\medskip
\textbf{Instructions:}
\begin{itemize}
  \item Read the [Scenario] and [Timelines] below.
  \item Choose the \textbf{one timeline} that best satisfies all convergent criteria.
  \item Reflect on why you chose that answer.
  \item Modify the $event\_description$ and the $timeline$ paths to make the question \textbf{harder}—aim to disguise the correct option such that it no longer feels obvious.
  \item Keep the number of timelines fixed at \{num\_options\}.
  \item Output the question in the same JSON format—no commentary, no extra explanation.
\end{itemize}

\textbf{Example}
\textbf{[Scenario]}  
It is January 2025. Donald Trump is re-elected and proposes a 20\% tariff on all foreign imports, including semiconductors and components. Nvidia relies heavily on Taiwan and South Korea for manufacturing, and sells to both U.S. and global cloud and enterprise clients. The proposal creates immediate uncertainty, but its implementation timeline is unclear.

\textbf{[Timelines]}  
\begin{enumerate}
  \item Tariff announcement → Nvidia pivots to cryptocurrency mining focus → Stock volatility rises → Nvidia delists from NASDAQ
  \item Tariff announcement → Immediate CapEx increase by Nvidia → Tariff-induced recession → Nvidia's gaming revenue rises → Stock soars
  \item Tariff announcement → Nvidia announces record profits → Stock surges 15\% → Tariffs boost investor confidence in U.S. production → Foreign competitors lose share
  \item Tariff announcement → Market volatility in semiconductor sector → Nvidia stock dips 10\% → Analysts revise gross margin outlook downward → Nvidia begins negotiating domestic manufacturing incentives → Medium-term uncertainty persists
\end{enumerate}

\textbf{[Answer]}  
4

\bigskip
Now apply this process to the following question:

\#\#\# Question to be Improved \#\#\#

\{question\}

\#\#\# End \#\#\#
\end{tcolorbox}

\subsection*{Convergent Thinking Answer Prompt}
\label{appendix:convergent-answer-prompt-template}
\begin{tcolorbox}[
  colback=gray!5!white,
  colframe=black!75!white,
  title=Convergent Thinking Model Answering Prompt,
  fonttitle=\bfseries,
  coltitle=white,
  arc=2mm,
  boxrule=0.4pt,
  breakable
]
You are a \textbf{300 IQ sell-side analyst} trained in \textbf{finance, economics, and politics}, who displays \textbf{top notch convergent thinking}, defined as the ability to assess whether a scenario or timeline:

\begin{enumerate}
  \item Exhibits strong \textbf{entailment} between each occurrence in the timeline,
  \item Has strong \textbf{factor alignment} with relevant economic/finance/political factors, and
  \item Shows strong \textbf{temporal coherence} between occurrences in the timeline.
\end{enumerate}

You are given a scenario and timeline options. Pick the timeline that most strongly satisfies the above convergent criteria.

\textbf{Instructions:}
\begin{itemize}
  \item Take a deep breath and think carefully about the scenario and the timelines provided.
  \item Reason through the scenario and the timelines.
  \item Choose the timeline that most strongly satisfies the convergent criteria.
  \item Output your thought process in the \texttt{<thinking>} section.
  \item Output your final answer in the \texttt{<answer>} section.
  \item Your response in the \texttt{<answer>} section must be a valid JSON string parsable by a JSON parser.
  \item Do not add any other text or preambles to the response.
  \item Do not leave a trailing comma after the last entry.
  \item \textbf{Give ONLY the JSON data.}
\end{itemize}

\textbf{Here is an example of the output format:}

\texttt{<thinking>}\\
\quad your reasoning steps\\
\texttt{</thinking>}\\

\texttt{<answer>}\\
\quad \{\\
\quad\quad "reason": \textless reason\_for\_answer\textgreater,\\
\quad\quad "final\_answer": \textless final\_answer\_as\_number\textgreater\\
\quad\}\\
\texttt{</answer>}

\vspace{5mm}
\textbf{This is the actual question:}

\textbf{[Scenario]}\\
Company: \{\texttt{company}\}\\
Event: \{\texttt{event}\}\\
Event Description: \{\texttt{event\_description}\}

\textbf{[Timelines]}\\
\{\texttt{timelines}\}

\textbf{[Output]}

\end{tcolorbox}

\subsection*{Convergent Thinking Error Analysis Prompt}
\label{appendix:convergent-error-analysis-prompt}
\begin{tcolorbox}[
  colback=gray!5!white,
  colframe=black!75!white,
  title=Convergent Thinking Model Answering Prompt,
  fonttitle=\bfseries,
  coltitle=white,
  arc=2mm,
  boxrule=0.4pt,
  breakable
]
You are an \textbf{error analysis assistant}. You will be given a multiple choice question, the model's answer, and the expected answer. 

You will analyze the model's answer and provide a \textbf{detailed explanation} of why the model's answer is incorrect. 

You will also provide a list of \textbf{possible reasons} for the model's incorrect answer. Please be as detailed as possible in your analysis.

Give a \textbf{category} for the error based on your reasons, and a \textbf{one-sentence summary} of the error.

\bigskip
\textbf{Output Format:}
\begin{verbatim}
<analysis>
{
    "model_answer": "<model_answer>",
    "expected_answer": "<expected_answer>",
    "possible_reasons": [
        "<possible_reason_1>",
        "<possible_reason_2>"
    ],
    "error_category": "<error_category>",
    "error_summary": "<error_summary>"
}
</analysis>
\end{verbatim}

\bigskip
\textbf{Input Fields:}

This is the question:

\texttt{\{question\}}

\medskip
\noindent\textbf{Model's Answer and Reasoning:}

\texttt{\{model\_answer\}}

\medskip
\noindent\textbf{Expected Answer and Reasoning:}

\texttt{\{expected\_answer\}}

\end{tcolorbox}

\section{Appendix IV - Convergent Thinking Errors}
\label{appendix:convergent-thinking errors}
\begin{longtable}{|p{0.95\textwidth}|}
\hline
\textbf{Examples of Model Evaluation Errors} \\
\hline
The model incorrectly selected Timeline (2) due to a misinterpretation of the scenario's focus on procedural delays and investor sentiment, which are better captured in Timeline (1). \\
\hline
The model misinterpreted the scenario's emphasis on compounded challenges leading to BRF's market exit and instead selected a timeline where BRF adapts and remains in the market. \\
\hline
The model incorrectly selected Timeline (2) due to a misinterpretation of the competitive and regulatory dynamics, failing to account for how Tenax's delay in expansion creates an opening for competitors as described in Timeline (4). \\
\hline
The model incorrectly selected Timeline (4) due to an overly optimistic interpretation of the scenario, failing to account for the realistic challenges and competitive dynamics emphasized in the expected answer. \\
\hline
The model incorrectly selected Timeline (2) due to a misinterpretation of the logical sequence of events and an overemphasis on temporal coherence, failing to align with the expected answer's reasoning. \\
\hline
The model misinterpreted the scenario by overemphasizing positive outcomes and failing to account for the logical entailment of subsidy uncertainty leading to dividend instability and prolonged stock decline. \\
\hline
The model incorrectly selected Timeline (3) due to a misinterpretation of the scenario's emphasis on financing challenges and their direct impact on the company's operational and financial outcomes. \\
\hline
The model incorrectly selected Timeline (3) due to a misinterpretation of causal relationships and an overemphasis on operational challenges, failing to prioritize the regulatory penalties and profit decline highlighted in Timeline (1). \\
\hline
The model incorrectly selected Timeline (3) due to a misinterpretation of the scenario's mixed signals and the company's strategic response, failing to recognize the stronger alignment of Timeline (4) with the expected optimism and international recovery focus. \\
\hline
The model incorrectly selected Timeline (2) due to a misinterpretation of the criteria, focusing on operational and financial balance rather than the alignment of consumer adoption and revenue growth emphasized in the expected answer. \\
\hline
The model incorrectly selected Timeline (2) due to a misinterpretation of the scenario, failing to account for the supply chain issues and competitive dynamics emphasized in the expected answer. \\
\hline
The model incorrectly selected Timeline (3) due to a misinterpretation of the competitive landscape and overemphasis on sustained growth, failing to account for the analysts' forecast and the importance of regional partnerships and digital wallet adoption in Timeline (2). \\
\hline
The model incorrectly selected Timeline (1) due to a failure to recognize the causal link between rising interest rates and increased demand for reverse factoring, which is central to Timeline (3). \\
\hline
The model incorrectly selected Timeline (3) due to a misinterpretation of the sequence of events and failure to prioritize derivative exposure concerns and acquisition news, which are central to Timeline (4). \\
\hline
The model misinterpreted the scenario details and overemphasized competitive pressures, leading to the selection of Timeline (3) instead of the correct Timeline (4), which better aligns with the scenario's steady demand and market dynamics. \\
\hline
The model incorrectly selected Timeline (2) due to an overemphasis on industry recognition and market sentiment, neglecting the scenario's focus on decentralized social media networks and strategic partnerships highlighted in Timeline (3). \\
\hline
The model incorrectly selected Timeline (3) due to an overly optimistic interpretation of the scenario, neglecting the emphasis on regulatory challenges, production delays, and competitive pressures that align more closely with Timeline (2). \\
\hline
The model incorrectly selected Timeline (3) by focusing on long-term strategic responses and growth potential, rather than the immediate challenges and negative outcomes described in Timeline (1). \\
\hline
The model's answer prioritizes broader impacts and temporal coherence over the expected answer's focus on factor alignment, particularly the consumer cost impact, leading to a misinterpretation of the correct timeline. \\
\hline
The model incorrectly selected Timeline (1) due to a misinterpretation of the scenario's emphasis on cyclical risks and geopolitical tensions, leading it to prioritize a timeline with positive outcomes over one that better reflects the expected challenges. \\
\hline
The model incorrectly selected Timeline (4) due to a misinterpretation of causal coherence and an overemphasis on innovation, failing to account for the stronger alignment of Timeline (3) with supply chain risks and competition factors. \\
\hline
The model incorrectly selected Timeline (1) by focusing on positive outcomes and temporal coherence, failing to account for the expected answer's emphasis on the complexity introduced by layoffs in Timeline (2). \\
\hline
The model incorrectly selected Timeline (3) due to a misinterpretation of the sequence of events and an overemphasis on bio-based adhesives, failing to recognize that Timeline (2) better aligns with regulatory adaptation and technological innovation. \\
\hline
The model incorrectly selected Timeline (3) due to a misinterpretation of the scenario and failure to recognize the logical alignment of Honeywell's exclusive focus on portable scanners in Timeline (1). \\
\hline
The model misinterpreted the scenario and selected a timeline that emphasizes operational strategies rather than the regulatory and financial consequences central to the expected answer. \\
\hline
The model incorrectly selected Timeline (4) due to an overemphasis on pricing strategy adjustments, overlooking the stronger strategic alignment and consumer-focused actions in Timeline (3). \\
\hline
\end{longtable}

\section{Appendix V: Intra-Model Correlations Across Divergent Dimensions}
\label{appendix: correlation-across-dimensions}
\begin{table}[ht]
\centering
\small
\begin{tabular}{lccccc}
\toprule
\textbf{} & \textbf{Plausibility} & \textbf{Novelty} & \textbf{Elaboration} & \textbf{Actionable} & \textbf{Richness} \\
\midrule
\textbf{Plausibility} & 1.000 & 0.706 & 0.496 & 0.576 & -0.043 \\
\textbf{Novelty}      & 0.706 & 1.000 & 0.581 & 0.540 & -0.002 \\
\textbf{Elaboration}  & 0.496 & 0.581 & 1.000 & 0.575 &  0.068 \\
\textbf{Actionable}   & 0.576 & 0.540 & 0.575 & 1.000 & -0.007 \\
\textbf{Richness}     & -0.043 & -0.002 & 0.068 & -0.007 & 1.000 \\
\bottomrule
\end{tabular}
\caption{Mean Average Correlation Across Models (Baseline)}
\label{tab:mean-correlation-baseline}
\end{table}

\begin{table}
\centering
\caption{phi4 Correlation Across Divergent Dimensions}
\label{tab:phi4_divergent_dim_corr}
\begin{tabular}{lrrrrr}
\toprule
{} &  Plausibility &  Novelty &  Elaboration &  Actionable &  Richness \\
\midrule
Plausibility &          1.00 &     0.87 &         0.80 &        0.68 &      0.02 \\
Novelty      &          0.87 &     1.00 &         0.78 &        0.66 &      0.01 \\
Elaboration  &          0.80 &     0.78 &         1.00 &        0.74 &     -0.00 \\
Actionable   &          0.68 &     0.66 &         0.74 &        1.00 &     -0.19 \\
Richness     &          0.02 &     0.01 &        -0.00 &       -0.19 &      1.00 \\
\bottomrule
\end{tabular}
\end{table}

\begin{table}
\centering
\caption{mistral\_lg Correlation Across Divergent Dimensions}
\label{tab:mistral_lg_divergent_dim_corr}
\begin{tabular}{lrrrrr}
\toprule
{} &  Plausibility &  Novelty &  Elaboration &  Actionable &  Richness \\
\midrule
Plausibility &          1.00 &     0.70 &         0.44 &        0.60 &     -0.01 \\
Novelty      &          0.70 &     1.00 &         0.61 &        0.69 &      0.02 \\
Elaboration  &          0.44 &     0.61 &         1.00 &        0.72 &      0.19 \\
Actionable   &          0.60 &     0.69 &         0.72 &        1.00 &      0.11 \\
Richness     &         -0.01 &     0.02 &         0.19 &        0.11 &      1.00 \\
\bottomrule
\end{tabular}
\end{table}

\begin{table}
\centering
\caption{llama3\_70b Correlation Across Divergent Dimensions}
\label{tab:llama3_70b_divergent_dim_corr}
\begin{tabular}{lrrrrr}
\toprule
{} &  Plausibility &  Novelty &  Elaboration &  Actionable &  Richness \\
\midrule
Plausibility &          1.00 &     0.78 &         0.61 &        0.61 &      0.03 \\
Novelty      &          0.78 &     1.00 &         0.51 &        0.58 &      0.06 \\
Elaboration  &          0.61 &     0.51 &         1.00 &        0.62 &      0.14 \\
Actionable   &          0.61 &     0.58 &         0.62 &        1.00 &      0.04 \\
Richness     &          0.03 &     0.06 &         0.14 &        0.04 &      1.00 \\
\bottomrule
\end{tabular}
\end{table}

\begin{table}[ht]
\centering
\small
\caption{o4\_mini Correlation Across Divergent Dimensions}
\label{tab:o4_mini_divergent_dim_corr}
\begin{tabular}{lccccc} 
\toprule
 & \textbf{Plausibility} & \textbf{Novelty} & \textbf{Elaboration} & \textbf{Actionable} & \textbf{Richness} \\
\midrule
\textbf{Plausibility} & 1.00 & 0.73 & 0.37 & 0.47 & -0.06 \\
\textbf{Novelty} & 0.73 & 1.00 & 0.60 & 0.51 & 0.02 \\
\textbf{Elaboration} & 0.37 & 0.60 & 1.00 & 0.55 & 0.19 \\
\textbf{Actionable} & 0.47 & 0.51 & 0.55 & 1.00 & 0.09 \\
\textbf{Richness} & -0.06 & 0.02 & 0.19 & 0.09 & 1.00 \\
\bottomrule
\end{tabular}
\end{table}

\begin{table}[ht]
\centering
\small
\caption{deepseek\_r1 Correlation Across Divergent Dimensions}
\label{tab:deepseek_r1_divergent_dim_corr}
\begin{tabular}{lccccc} 
\toprule
 & \textbf{Plausibility} & \textbf{Novelty} & \textbf{Elaboration} & \textbf{Actionable} & \textbf{Richness} \\
\midrule
\textbf{Plausibility} & 1.00 & -0.16 & 0.22 & 0.60 & -0.10 \\
\textbf{Novelty} & -0.16 & 1.00 & 0.39 & 0.08 & -0.04 \\
\textbf{Elaboration} & 0.22 & 0.39 & 1.00 & 0.31 & -0.03 \\
\textbf{Actionable} & 0.60 & 0.08 & 0.31 & 1.00 & -0.06 \\
\textbf{Richness} & -0.10 & -0.04 & -0.03 & -0.06 & 1.00 \\
\bottomrule
\end{tabular}
\end{table}

\begin{table}[ht]
\centering
\small
\caption{llama3\_8b Correlation Across Divergent Dimensions}
\label{tab:llama3_8b_divergent_dim_corr}
\begin{tabular}{lccccc} 
\toprule
 & \textbf{Plausibility} & \textbf{Novelty} & \textbf{Elaboration} & \textbf{Actionable} & \textbf{Richness} \\
\midrule
\textbf{Plausibility} & 1.00 & 0.78 & 0.48 & 0.46 & -0.13 \\
\textbf{Novelty} & 0.78 & 1.00 & 0.34 & 0.40 & -0.17 \\
\textbf{Elaboration} & 0.48 & 0.34 & 1.00 & 0.54 & -0.03 \\
\textbf{Actionable} & 0.46 & 0.40 & 0.54 & 1.00 & -0.03 \\
\textbf{Richness} & -0.13 & -0.17 & -0.03 & -0.03 & 1.00 \\
\bottomrule
\end{tabular}
\end{table}

\begin{table}[ht]
\centering
\small
\caption{gpt4o Correlation Across Divergent Dimensions}
\label{tab:gpt4o_divergent_dim_corr}
\begin{tabular}{lccccc}
\toprule
 & \textbf{Plausibility} & \textbf{Novelty} & \textbf{Elaboration} & \textbf{Actionable} & \textbf{Richness} \\
\midrule
\textbf{Plausibility} & 1.00 & 0.65 & 0.35 & 0.51 & -0.02 \\
\textbf{Novelty} & 0.65 & 1.00 & 0.80 & 0.61 & 0.08 \\
\textbf{Elaboration} & 0.35 & 0.80 & 1.00 & 0.58 & 0.12 \\
\textbf{Actionable} & 0.51 & 0.61 & 0.58 & 1.00 & 0.08 \\
\textbf{Richness} & -0.02 & 0.08 & 0.12 & 0.08 & 1.00 \\
\bottomrule
\end{tabular}
\end{table}

\begin{table}[ht]
\centering
\small
\caption{mistral\_sm Correlation Across Divergent Dimensions}
\label{tab:mistral_sm_divergent_dim_corr}
\begin{tabular}{lccccc}
\toprule
 & \textbf{Plausibility} & \textbf{Novelty} & \textbf{Elaboration} & \textbf{Actionable} & \textbf{Richness} \\
\midrule
\textbf{Plausibility} & 1.00 & 0.87 & 0.69 & 0.72 & -0.17 \\
\textbf{Novelty} & 0.87 & 1.00 & 0.79 & 0.77 & -0.12 \\
\textbf{Elaboration} & 0.69 & 0.79 & 1.00 & 0.74 & -0.10 \\
\textbf{Actionable} & 0.72 & 0.77 & 0.74 & 1.00 & -0.16 \\
\textbf{Richness} & -0.17 & -0.12 & -0.10 & -0.16 & 1.00 \\
\bottomrule
\end{tabular}
\end{table}

\begin{table}[ht]
\centering
\small
\caption{llama4\_scout Correlation Across Divergent Dimensions}
\label{tab:llama4_scout_divergent_dim_corr}
\begin{tabular}{lccccc}
\toprule
 & \textbf{Plausibility} & \textbf{Novelty} & \textbf{Elaboration} & \textbf{Actionable} & \textbf{Richness} \\
\midrule
\textbf{Plausibility} & 1.00 & 0.70 & 0.70 & 0.59 & 0.02 \\
\textbf{Novelty} & 0.70 & 1.00 & 0.51 & 0.58 & 0.13 \\
\textbf{Elaboration} & 0.70 & 0.51 & 1.00 & 0.66 & 0.13 \\
\textbf{Actionable} & 0.59 & 0.58 & 0.66 & 1.00 & 0.11 \\
\textbf{Richness} & 0.02 & 0.13 & 0.13 & 0.11 & 1.00 \\
\bottomrule
\end{tabular}
\end{table}

\begin{table}[ht]
\centering
\small
\caption{cohere\_command\_r Correlation Across Divergent Dimensions}
\label{tab:cohere_command_r_divergent_dim_corr}
\begin{tabular}{lccccc}
\toprule
 & \textbf{Plausibility} & \textbf{Novelty} & \textbf{Elaboration} & \textbf{Actionable} & \textbf{Richness} \\
\midrule
\textbf{Plausibility} & 1.00 & 0.79 & 0.54 & 0.70 & -0.03 \\
\textbf{Novelty} & 0.79 & 1.00 & 0.61 & 0.71 & 0.06 \\
\textbf{Elaboration} & 0.54 & 0.61 & 1.00 & 0.80 & 0.12 \\
\textbf{Actionable} & 0.70 & 0.71 & 0.80 & 1.00 & 0.04 \\
\textbf{Richness} & -0.03 & 0.06 & 0.12 & 0.04 & 1.00 \\
\bottomrule
\end{tabular}
\end{table}

\begin{table}[ht]
\centering
\small
\caption{cohere\_command\_a Correlation Across Divergent Dimensions}
\label{tab:cohere_command_a_divergent_dim_corr}
\begin{tabular}{lccccc}
\toprule
 & \textbf{Plausibility} & \textbf{Novelty} & \textbf{Elaboration} & \textbf{Actionable} & \textbf{Richness} \\
\midrule
\textbf{Plausibility} & 1.00 & 0.86 & 0.33 & 0.51 & -0.22 \\
\textbf{Novelty} & 0.86 & 1.00 & 0.58 & 0.51 & -0.26 \\
\textbf{Elaboration} & 0.33 & 0.58 & 1.00 & 0.35 & -0.16 \\
\textbf{Actionable} & 0.51 & 0.51 & 0.35 & 1.00 & -0.26 \\
\textbf{Richness} & -0.22 & -0.26 & -0.16 & -0.26 & 1.00 \\
\bottomrule
\end{tabular}
\end{table}

\begin{table}[ht]
\centering
\small
\caption{o1 Correlation Across Divergent Dimensions}
\label{tab:o1_divergent_dim_corr}
\begin{tabular}{lccccc}
\toprule
 & \textbf{Plausibility} & \textbf{Novelty} & \textbf{Elaboration} & \textbf{Actionable} & \textbf{Richness} \\
\midrule
\textbf{Plausibility} & 1.00 & 0.84 & 0.28 & 0.60 & -0.03 \\
\textbf{Novelty} & 0.84 & 1.00 & 0.51 & 0.49 & 0.01 \\
\textbf{Elaboration} & 0.28 & 0.51 & 1.00 & 0.29 & 0.08 \\
\textbf{Actionable} & 0.60 & 0.49 & 0.29 & 1.00 & -0.04 \\
\textbf{Richness} & -0.03 & 0.01 & 0.08 & -0.04 & 1.00 \\
\bottomrule
\end{tabular}
\end{table}

\begin{table}[ht]
\centering
\small
\caption{llama4\_maverick Correlation Across Divergent Dimensions}
\label{tab:llama4_maverick_divergent_dim_corr}
\begin{tabular}{lccccc}
\toprule
 & \textbf{Plausibility} & \textbf{Novelty} & \textbf{Elaboration} & \textbf{Actionable} & \textbf{Richness} \\
\midrule
\textbf{Plausibility} & 1.00 & 0.73 & 0.74 & 0.54 & 0.07 \\
\textbf{Novelty} & 0.73 & 1.00 & 0.47 & 0.47 & 0.05 \\
\textbf{Elaboration} & 0.74 & 0.47 & 1.00 & 0.60 & 0.07 \\
\textbf{Actionable} & 0.54 & 0.47 & 0.60 & 1.00 & 0.08 \\
\textbf{Richness} & 0.07 & 0.05 & 0.07 & 0.08 & 1.00 \\
\bottomrule
\end{tabular}
\end{table}

\begin{table}[ht]
\centering
\small
\caption{o1\_mini Correlation Across Divergent Dimensions}
\label{tab:o1_mini_divergent_dim_corr}
\begin{tabular}{lccccc}
\toprule
 & \textbf{Plausibility} & \textbf{Novelty} & \textbf{Elaboration} & \textbf{Actionable} & \textbf{Richness} \\
\midrule
\textbf{Plausibility} & 1.00 & 0.74 & 0.40 & 0.47 & 0.04 \\
\textbf{Novelty} & 0.74 & 1.00 & 0.62 & 0.49 & 0.12 \\
\textbf{Elaboration} & 0.40 & 0.62 & 1.00 & 0.56 & 0.24 \\
\textbf{Actionable} & 0.47 & 0.49 & 0.56 & 1.00 & 0.10 \\
\textbf{Richness} & 0.04 & 0.12 & 0.24 & 0.10 & 1.00 \\
\bottomrule
\end{tabular}
\end{table}

\clearpage
\section{Appendix VI: Model Call Configuration}
The following parameters were used in the invocation of the language model through the `chat\_completion` function:

\begin{itemize}[leftmargin=2em]
  \item \textbf{System Prompt:} A system instruction string guiding the assistant's behavior.
  \item \textbf{User Message:} The primary user input passed to the assistant.
  \item \textbf{Model:} The model name is specified through the \texttt{config.model\_name} field.
  \item \textbf{max\_tokens:} \texttt{4096} \\
  This parameter sets the maximum number of tokens to generate in the response.
  \item \textbf{temperature:} \texttt{1.0} \\
  Controls randomness in generation. Higher values produce more diverse outputs.
  \item \textbf{top\_p:} \texttt{1.0} \\
  Nucleus sampling parameter. Limits token selection to a subset with cumulative probability \texttt{top\_p}.
  \item \textbf{response\_format:} \texttt{json\_object} \\
  Responses are structured in JSON format to enforce schema compliance.
\end{itemize}

Parameters above do not apply to reasoning models such as o1, o4\_mini, and deepseek\_r1, which are called with default arguments.


\begin{thebibliography}{30}

\bibitem{srivastava2023imitationgame}
Aarohi Srivastava, Abhinav Rastogi, Abhishek Rao, Abu Awal Md Shoeb, Abubakar Abid, Adam Fisch, Adam R. Brown, Adam Santoro, Aditya Gupta, Adrià Garriga-Alonso, et al.
\textit{Beyond the Imitation Game: Quantifying and extrapolating the capabilities of language models}.
Transactions on Machine Learning Research, 2023.
\url{https://arxiv.org/abs/2206.04615}.

\bibitem{hendrycks2021mmlu}
Dan Hendrycks, Collin Burns, Steven Basart, Andy Zou, Mantas Mazeika, Dawn Song, and Jacob Steinhardt.
\newblock Measuring Massive Multitask Language Understanding.
\newblock In \emph{Proceedings of the International Conference on Learning Representations (ICLR)}, 2021.
\newblock \url{https://arxiv.org/abs/2009.03300}.

\bibitem{clark2018arc}
Peter Clark, Isaac Cowhey, Oren Etzioni, Tushar Khot, Ashish Sabharwal, Carissa Schoenick, and Oyvind Tafjord.
\newblock Think you have Solved Question Answering? Try ARC, the AI2 Reasoning Challenge.
\newblock In \emph{Proceedings of the 2018 Conference on Empirical Methods in Natural Language Processing (EMNLP)}, 2018.
\newblock \url{https://arxiv.org/abs/1803.05457}.

\bibitem{mihaylov2018openbookqa}
Todor Mihaylov, Peter Clark, Tushar Khot, and Ashish Sabharwal.
\newblock Can a Suit of Armor Conduct Electricity? A New Dataset for Open Book Question Answering.
\newblock In \emph{Proceedings of the 2018 Conference on Empirical Methods in Natural Language Processing (EMNLP)}, pages 2381–2391, Brussels, Belgium, 2018.
\newblock Association for Computational Linguistics.
\newblock \url{https://aclanthology.org/D18-1260/}.

\bibitem{bellemare2024divergent}
Antoine Bellemare-Pepin, François Lespinasse, Philipp Thölke, Yann Harel, Kory Mathewson, Jay A. Olson, Yoshua Bengio, and Karim Jerbi.
\newblock Divergent Creativity in Humans and Large Language Models.
\newblock \emph{arXiv preprint arXiv:2405.13012}, 2024.

\bibitem{chen2023theoremqa}
Wenhu Chen, Ming Yin, Max Ku, Pan Lu, Yixin Wan, Xueguang Ma, Jianyu Xu, Xinyi Wang, and Tony Xia.
\newblock TheoremQA: A Theorem-driven Question Answering Dataset.
\newblock In \emph{Proceedings of the 2023 Conference on Empirical Methods in Natural Language Processing (EMNLP)}, pages 7889–7901. Association for Computational Linguistics, Singapore, 2023.

\bibitem{golde2025mastermind}
Jonas Golde, Patrick Haller, Fabio Barth, and Alan Akbik.
\newblock MASTERMINDEVAL: A Simple but Scalable Reasoning Benchmark.
\newblock \emph{arXiv preprint arXiv:2503.05891}, 2025. (To be published at ICLR 2025 Workshop on Reasoning and Planning for LLMs).

\bibitem{ho2020constructing}
Xanh Ho, Anh-Khoa Duong Nguyen, Saku Sugawara, and Akiko Aizawa.
\newblock Constructing A Multi-hop QA Dataset for Comprehensive Evaluation of Reasoning Steps.
\newblock In \emph{Proceedings of the 28th International Conference on Computational Linguistics (COLING)}, pages 6609–6625. International Committee on Computational Linguistics, Barcelona, Spain (Online), 2020.

\bibitem{kumar2025human}
Harsh Kumar, Jonathan Vincentius, Ewan Jordan, and Ashton Anderson.
\newblock Human Creativity in the Age of LLMs: Randomized Experiments on Divergent and Convergent Thinking.
\newblock In \emph{Proceedings of the CHI Conference on Human Factors in Computing Systems (CHI '25)}. ACM, New York, NY, USA, 2025. DOI:\url{https://doi.org/10.1145/3706598.3714198}.

\bibitem{kwiatkowski2019natural}
Tom Kwiatkowski, Jennimaria Palomaki, Olivia Redfield, Michael Collins, Ankur Parikh, Chris Alberti, Danielle Epstein, Illia Polosukhin, Jacob Devlin, Kenton Lee, Kristina Toutanova, Llion Jones, Matthew Kelcey, Ming-Wei Chang, Andrew M. Dai, Jakob Uszkoreit, Quoc Le, and Slav Petrov.
\newblock Natural Questions: A Benchmark for Question Answering Research.
\newblock \emph{Transactions of the Association for Computational Linguistics}, 7:452–466, 2019.

\bibitem{liang2024encouraging}
Tian Liang, Zhiwei He, Wenxiang Jiao, Xing Wang, Yan Wang, Rui Wang, Yujiu Yang, Shuming Shi, and Zhaopeng Tu.
\newblock Encouraging Divergent Thinking in Large Language Models through Multi-Agent Debate.
\newblock \emph{arXiv preprint arXiv:2305.19118}, 2024.

\bibitem{lu2024benchmarking}
Yining Lu, Dixuan Wang, Tianjian Li, Dongwei Jiang, and Daniel Khashabi.
\newblock Benchmarking Language Model Creativity: A Case Study on Code Generation.
\newblock \emph{arXiv preprint arXiv:2407.09007}, 2024.

\bibitem{mirzadeh2024gsm}
Iman Mirzadeh, Keivan Alizadeh, Hooman Shahrokhi, Oncel Tuzel, Samy Bengio, and Mehrdad Farajtabar.
\newblock GSM-Symbolic: Understanding the Limitations of Mathematical Reasoning in Large Language Models.
\newblock \emph{arXiv preprint arXiv:2410.05229}, 2024.

\bibitem{nguyen2016msmarco}
Tri Nguyen, Mir Rosenberg, Xia Song, Jianfeng Gao, Saurabh Tiwary, Rangan Majumder, and Li Deng.
\newblock MS MARCO: A Human Generated MAchine Reading COmprehension Dataset.
\newblock \emph{arXiv preprint arXiv:1611.09268}, 2016.

\bibitem{rein2023gpqa}
David Rein, Betty Li Hou, Asa Cooper Stickland, Jackson Petty, Richard Yuanzhe Pang, Julien Dirani, Julian Michael, and Samuel R. Bowman.
\newblock GPQA: A Graduate-Level Google-Proof Q\&A Benchmark.
\newblock \emph{arXiv preprint arXiv:2311.12022}, 2023.

\bibitem{sawada2023arb}
Tomohiro Sawada, Daniel Paleka, Alexander Havrilla, Pranav Tadepalli, Paula Vidas, Alexander Kranias, John J. Nay, Kshitij Gupta, and Aran Komatsuzaki.
\newblock ARB: Advanced Reasoning Benchmark for Large Language Models.
\newblock \emph{arXiv preprint arXiv:2307.13692}, 2023.

\bibitem{su2025bright}
Hongjin Su, Howard Yen, Mengzhou Xia, Weijia Shi, Niklas Muennighoff, Han-yu Wang, Haisu Liu, Quan Shi, Zachary S. Siegel, Michael Tang, Ruoxi Sun, Jinsung Yoon, Sercan Ö. Arık, Danqi Chen, and Tao Yu.
\newblock BRIGHT: A Realistic and Challenging Benchmark for Reasoning-Intensive Retrieval.
\newblock In \emph{Proceedings of the International Conference on Learning Representations (ICLR 2025)}, 2025.

\bibitem{thakur2021beir}
Nandan Thakur, Nils Reimers, Andreas Rücklé, Abhishek Srivastava, and Iryna Gurevych.
\newblock BEIR: A Heterogeneous Benchmark for Zero-shot Evaluation of Information Retrieval Models.
\newblock In \emph{Proceedings of the Neural Information Processing Systems Track on Datasets and Benchmarks}, 2021.

\bibitem{wu2024mrke}
Jian Wu, Linyi Yang, Manabu Okumura, and Yue Zhang.
\newblock MRKE: The Multi-hop Reasoning Evaluation of LLMs by Knowledge Edition.
\newblock \emph{arXiv preprint arXiv:2402.11924}, 2024.

\bibitem{wu2024cofca}
Jian Wu, Linyi Yang, Zhen Wang, Manabu Okumura, and Yue Zhang.
\newblock COFCA: A Step-wise Counterfactual Multi-hop QA Benchmark.
\newblock \emph{arXiv preprint arXiv:2402.11924}, 2024.

\bibitem{yang2018hotpotqa}
Zhilin Yang, Peng Qi, Saizheng Zhang, Yoshua Bengio, William W. Cohen, Ruslan Salakhutdinov, and Christopher D. Manning.
\newblock HotpotQA: A Dataset for Diverse, Explainable Multi-hop Question Answering.
\newblock In \emph{Proceedings of the 2018 Conference on Empirical Methods in Natural Language Processing (EMNLP)}, pages 2369–2380. Association for Computational Linguistics, Brussels, Belgium, 2018.

\bibitem{yuan2025naturalreasoning}
Weizhe Yuan, Jane Yu, Song Jiang, Karthik Padthe, Yang Li, Ilia Kulikov, Kyunghyun Cho, Dong Wang, Yuandong Tian, Jason Weston, and Xian Li.
\newblock NATURALREASONING: Reasoning in the Wild with 2.8M Challenging Questions.
\newblock \emph{arXiv preprint arXiv:2502.13124}, 2025.

\bibitem{renze2024selfreflection}
M. Renze and E. Guven.
\newblock Self-reflection in LLM Agents: Effects on Problem-solving Performance.
\newblock \emph{arXiv preprint arXiv:2405.06682}, 2024.

\bibitem{lampinen2022explanations}
Andrew K. Lampinen, Ishita Dasgupta, Stephanie C. Y. Chan, Kory Matthewson, Michael Henry Tessler, Antonia Creswell, James L. McClelland, Jane X. Wang, and Felix Hill.
\newblock Can Language Models Learn from Explanations in Context?
\newblock \emph{arXiv preprint arXiv:2204.02329}, 2022.

\end{thebibliography}
\end{document}